\documentclass[preprint,12pt]{elsarticle}
\makeatletter
\def\ps@pprintTitle{%
 \let\@oddhead\@empty
 \let\@evenhead\@empty
 \def\@oddfoot{}%
 \def\@evenfoot{}%
}
\makeatother

\usepackage{amssymb}
\usepackage{amsmath}
\usepackage{latexsym}

\usepackage[switch]{lineno}

\usepackage{caption}
\usepackage{subcaption}

\usepackage{bm}

\usepackage{siunitx}

\usepackage{romannum}

\usepackage{float}

\usepackage{url}
\usepackage{xcolor}
\definecolor{newcolor}{rgb}{.8,.349,.1}

\usepackage[citebordercolor=white]{hyperref}
\usepackage{cleveref}

\usepackage{appendix}
\journal{Acta Astronautica}

\begin{document}

\begin{frontmatter}



\title{Visualizing Critic Match Loss Landscapes for Interpretation of Online Reinforcement Learning Control Algorithms}%


% \author{} %% Author name

% \affiliation{organization={},%Department and Organization

\author[1]{Jingyi Liu}
\author[1]{Jian Guo\corref{cor1}}
\cortext[cor1]{Corresponding author} 
\ead{J.Guo@tudelft.nl}
\author[1]{Eberhard Gill}

%\address[2]{Affiliation 1, Address, City and Postal Code, Country}
\affiliation[1]{organization={Faculty of Aerospace Engineering, Delft University of Technology},
                addressline={Kluyverweg 1},
                city={Delft},
                postcode={2629 HS},
                country={The Netherlands}}

%% Abstract
\begin{abstract}
Reinforcement learning has proven its power on various occasions. However, its performance is not always guaranteed when system dynamics change. Instead, it largely relies on users' empirical experience. For reinforcement learning algorithms with an actor-critic structure, the critic neural network reflects the approximation and optimization process in the RL algorithm. Analyzing the performance of the critic neural network helps to understand the mechanism of the algorithm. To support systematic interpretation of such algorithms in dynamic control problems, this work proposes a critic match loss landscape visualization method for online reinforcement learning.{The method constructs a loss landscape by projecting recorded critic parameter trajectories onto a low-dimensional linear subspace. The critic match loss is evaluated over the projected parameter grid using fixed reference state samples and temporal-difference targets. This yields a three-dimensional loss surface together with a two-dimensional optimization path that characterizes critic learning behavior. To extend analysis beyond visual inspection, quantitative landscape indices and a normalized system performance index are introduced, enabling structured comparison across different training outcomes. The approach is demonstrated using the Action-Dependent Heuristic Dynamic Programming algorithm on cart-pole and spacecraft attitude control tasks. Comparative analyses across projection methods and training stages reveal distinct landscape characteristics associated with stable convergence and unstable learning. The proposed framework enables both qualitative and quantitative interpretation of critic optimization behavior in online reinforcement learning.}
\end{abstract}


% \begin{highlights}
% \end{highlights}

%% Keywords

\end{frontmatter}

\clearpage
\pagenumbering{arabic}


%% For linenumbers
%%\linenumbers

\section{Introduction}
\label{sec1}
% Please use \verb+elsarticle.cls+ for typesetting your paper. Additionally,
% make sure not to remove the package \verb+jasr.sty+ already included in the
% preamble:
% \begin{verbatim} 
%   \usepackage{jasr}
% \end{verbatim}

% Make sure to have the file \verb+jasr-model5-names.bst+ to produce the references in
% the correct format. 

% Any instructions relevant to the \verb+elsarticle.cls+ are
% applicable here as well. See the online instruction available at:
% \makeatletter
% \if@twocolumn
% \begin{verbatim}
%  https://support.stmdocs.in/wiki/
%  index.php?title=Elsarticle.cls
% \end{verbatim}
% \else
% \begin{verbatim}
%  https://support.stmdocs.in/wiki/index.php?title=Elsarticle.cls
%  \end{verbatim}
% \fi
% \makeatother

% Following commands are defined for this journal which are not in
% \verb+elsarticle.cls+. 
% \begin{verbatim}
%   \received{}
%   \finalform{}
%   \accepted{}
%   \availableonline{}
%   \communicated{}
% \end{verbatim}

Reinforcement learning algorithms have been a research hot spot in recent years. These algorithms have demonstrated impressive performance in areas such as robotics~\citep{ibarz2021train}, game playing~\citep{silver2017mastering}, navigation and control~\citep{drones6100270}, and decision-making tasks~\citep{lee2012neural}.

For system control with uncertainties, reinforcement learning methods have also been widely applied~\citep{duan2016benchmarking}. For example, in Active Debris Removal (ADR) using robotic arms, the system becomes a new combined spacecraft system with uncertainties due to the uncooperative target capture. For combined spacecraft with system uncertainties, several works have been using reinforcement learning methods ~\citep{oestreich2021autonomous,tipaldi2022reinforcement, rafiee2024active}, which rely on the sampled data during the control process, rather than on model information.

In some control scenarios, the environment changes constantly. For example, in the above-mentioned ADR scenario, the combined spacecraft system can be an uncertain system with an unknown moving appendage, which causes the system dynamics to constantly change. As a result, a real-time RL agent that interacts with the environment is needed. This is where online reinforcement learning is applied. For example, the Adaptive Dynamic Programming (ADP) method is originally a method based on optimal control. With the actor-critic (AC) architecture, it can be regarded as one kind of online reinforcement learning method.  Action-Dependent Heuristic Dynamic Programming (ADHDP), or Q learning, was introduced as a supplementary controller to guarantee a proper dynamic performance \citep{wei2018learning}. ADHDP is also used as the sole controller in the post-capture scenario with unknown system parameters \citep{Gao2019thesis}. 

 Despite the successful applications of reinforcement learning control mentioned above,  there are also possibilities that reinforcement learning algorithms do not perform as expected. Since RL algorithms are commonly trained on a specific fixed environment, it may work for one system, yet it fails to generalize to another \citep{packer2018assessing}. In the extreme condition, when a single system parameter changes, a well-tuned RL algorithm may fail to work \citep{peng2018sim}. Therefore, the interpretation of the algorithm's performance is important for understanding the mechanism of the algorithm and for improving the performance of the algorithm.

 The interpretation of the reinforcement learning algorithm largely relies on visualization techniques. {Visualization of learning processes and control performance is one aspect for interpretation of RL algorithms. In many learning-based control studies, classical visualization techniques are adopted, such as learning curves, parameter evolution, and system performance trajectories, to illustrate convergence behavior and control effectiveness \citep{10065569,10163245}. To further support the interpretation of reinforcement learning behavior, visualization techniques have been developed to examine loss functions and optimization landscapes.} Using visualization methods based on "filter normalization", the structure of neural loss functions is explored and the effect of loss landscapes on generalization\citep{li2018visualizing}.  The filter-normalized method is further used to visualize reward surfaces for popular reinforcement learning environments \citep{sullivan2022cliff}. By mapping between a policy and return, the return landscape is generated, which can navigate away from noisy neighborhoods and thus improve policy robustness \citep{rahn2023policy}. The characteristics of the actor loss function are studied by visualizing the loss functions \citep{bekci2020visualizing}. By visualizing the optimization surface of the Reacher and Walker-walk task implemented by Proximal Policy Optimization (PPO), the action representation's significant influence on the learning performance is demonstrated \citep{schneider2023investigating}.

 {The above-mentioned works mainly use visualization techniques to explore reward landscapes and actor loss landscapes, and interpret algorithm performance from the perspectives of policy outcome or actor optimization. Reward surfaces describe how changes in policy parameters affect cumulative return, while actor loss landscapes reflect characteristics of policy update mechanisms. However, these visualizations do not directly reveal how the critic module in the actor–critic structure is optimized, nor do they visualize the geometry of the critic module. In the actor–critic structure, the critic module is applied to approximate the value function or cost function, and its approximation accuracy significantly influences or even governs the learning stability of the algorithm. Visualizing the critic behavior, therefore, helps to reveal the learning mechanism of the critic module and provides useful information for the interpretation of reinforcement learning algorithms.}

In reinforcement learning with an actor–critic structure, the critic network is trained by minimizing the temporal-difference (TD) error step by step, and the approximation process is carried out through parameter updates. The approximation accuracy of the critic largely impacts the convergence behavior of the algorithm. The training of a critic network is a process to minimize the cost by updating the weights, i.e., the parameters used for approximation. During online training, both the TD target and the state distribution evolve as the policy changes, which makes the TD objective inherently moving and difficult to visualize as a single and well-defined surface. {To enable interpretation of critic learning behavior under such conditions, we construct a critic match loss derived from the TD error by fixing the reference data and targets associated with a given policy. Since the TD error is the training signal that directly drives the critic parameter updates, the resulting critic match loss provides a meaningful representation of critic learning behavior that can be visualized as a well-defined loss landscape, which allows the critic optimization process to be interpreted from a geometric perspective.}

In this work, we visualize the performance of the critic neural network for online reinforcement learning by constructing a critic match loss landscape. The critic weight parameters at the end of each training episode are recorded to keep track of the optimization path of critic weight during training. The corresponding state data and temporal-difference targets associated with selected policy references are used to reconstruct the critic match loss landscape over candidate weight parameters. These parameters are projected onto a principal component plane based on the optimization path of critic weight. Using the method above, the derived 3-D loss landscape and 2-D loss path can qualitatively reveal the training evolution and expose local geometry \textbf{blue}{under a given policy reference}. It can also help to explain why a single RL algorithm converges in one system yet diverges in another.
  
The remainder of this paper is organized as follows. In Section 2, the critic loss landscape visualization method for online RL is introduced. The ADHDP algorithm is presented as a sample algorithm for the visualization technique. {The quantitative indices for analyzing the loss landscape and system performance are given.} In Section 3, using the cart-pole system and the spacecraft attitude system as the control object, the dynamic models of the two systems are introduced. The corresponding control results using ADHDP algorithms are shown. In Section 4, {the performance of the ADHDP algorithm is interpreted using the critic loss landscape visualization method. Comparisons are made across systems, across projection methods and selected policies.} In Section 5, a conclusion is drawn.

\section{Method}

In this section, the visualization method for interpreting online RL algorithms is introduced. The ADHDP algorithm is also given, which is used as an object to be interpreted using the visualization method. {Quantitative indices for quantitatively analyzing the loss landscapes with system performance are given.}

\subsection{Visualizing the loss function}
Reinforcement learning uses neural networks as a tool to approximate functions. Normally, these neural networks approximate complex functions with many parameters. Visualizing the loss with these parameters can explicitly reveal the characteristics of the neural net. For example, the flatness and stiffness of the neural net, local minima, and saddle points. All of these interesting characteristics will be visible in the loss landscape. However, the large number of parameters makes the loss landscape visualization a multi-dimensional problem, which is difficult to visualize. To tackle this problem, the work \citep{li2018visualizing} used a method called Contour Plots and Random Directions. The main idea of the method is that, since it's not possible to fully visualize how the loss changes with the full-dimensional parameters, two dimensions are selected to generate a 3-D loss landscape. The 3-D loss landscape is a projection slice of the full-dimensional loss landscape. To calculate the loss regarding the two selected dimensions, the following equation is used 

\begin{equation}
f(\alpha, \beta) = L(\theta^* + \alpha \boldsymbol{\delta} + \beta \boldsymbol{\eta})
\label{eq: contour Plots}
\end{equation}

Here, $\theta$ is the chosen center point in the landscape graph while $\boldsymbol{\delta}$ and $\boldsymbol{\eta}$ are the two directions chosen to visualize the landscape. The parameters $\alpha$ and $\beta$ are the distances from the chosen center point.

The equation could be understood in the following way. The multidimensional problem is reduced to a 3-D coordinate system. The loss is visualized on a 2-D subspace spanned by the chosen directions $\boldsymbol{\delta}$ and $\boldsymbol{\eta}$. The 3-D plot shows the loss value $f$ along the vertical axis, with $\alpha$ and $\beta$ as coordinates on the horizontal plane.

Using the aforementioned Contour Plots and Random Direction method, the multidimensional problem has been reduced to a 3-D  problem, which is more explicit and straightforward to visualize. 

\subsection{Action-dependent heuristic dynamic programming (ADHDP)}

\begin{figure}[H]
\centering
\includegraphics[width = 0.7\textwidth]{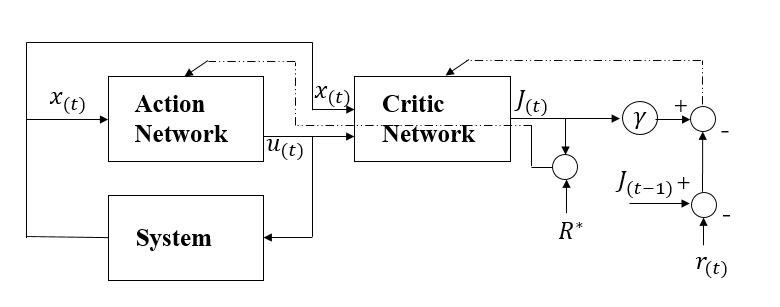}
\caption{Structure of ADHDP}
\label{fig:ADHDP_structure}
\end{figure}

To illustrate how the critic loss landscape visualization method contributes to the interpretation of online RL methods, Action-dependent heuristic dynamic programming (ADHDP), a specific RL method, is used here as an example. ADHDP is a specific type of reinforcement learning method. In this method, the objectives of controlling and learning are separated and realized with two separate neural networks, which are the actor neural network and the critic neural network, as shown in ~\autoref{fig:ADHDP_structure} \citep{si2004handbook}. The critic network is trained toward optimizing the total reward-to-go objective, which is the approximation of the cost function. The actor neural network is trained so that the critic network generates an ultimate objective of success, which is to minimize the cost. 

In ADHDP, the cost function $J(t)$ is expressed as
\begin{eqnarray}
J_{(t)}=r(t+1)+\gamma r(t+2)+\cdots=\sum_{k=1}^{\infty} \gamma^{k-1} r(t+k)
\label{eq:discounted cost function}
\end{eqnarray}

Here, ${r(t+1)}$ is the defined reinforcement signal or reward for the system. $k$ is the number of time steps from time instance $t$. The expression of the signal equation can be defined according to the goal of control. The discount factor $\gamma$ indicates how much rewards in the distant future can influence those in the immediate future.

We define the prediction error $e_c$ for the critic network as 
\begin{eqnarray}
e_c(t)=[r(t)+\gamma J(t)]-J(t-1)
\label{eq:TD Error}
\end{eqnarray}
The prediction error is also called TD error in reinforcement learning. 

The set of weight parameters in the critic network is $\mathbf{w}_c
$, as shown in \autoref {eq:cost function approx}. The weight update rule for the critic network is based on the gradient descent method. 

In this paper, the critic and actor networks are approximated using the Multilayer Perceptron (MLP) structure with one hidden layer. Detained information about the ADHDP algorithm can be found in ~\ref{Appendix}.

\subsection{Visualizing the critic match loss function for ADHDP}
\label{subsection: ADHDP critic match loss method}
The ADHDP algorithm is trained based on 
 \autoref{eq:TD Error}.The approximation precision of cost function J will influence the performance of the algorithm. The approximation error, TD error in  \autoref{eq:TD Error} is the loss for the critic network. Hence, visualizing the critic match loss landscape will reveal the approximation process of the cost function. From the landscape, we can observe the general trend of how the algorithm evolves during its update.

The critic network is implemented as a multilayer perceptron (MLP) with parameters denoted by $\mathbf{w}_c$. These parameters are updated at each time step, reflecting changes in the critic network during training. During online training the system is reset at the beginning of each episode while the network parameters continue from the previous episode. To track this evolution, the vector \( \mathbf{w}_c \) is recorded at the end of each episode $k$, forming a sequence \( [\mathbf{w}_c(k), \mathbf{w}_c(k+1), \dots, \mathbf{w}_c(k+n)] \).

From \autoref{eq: contour Plots}, two directions have to be selected to function as the two axes for landscape visualization. To select the two directions, the Principal Component Analysis (PCA) method is applied to generate the two orthogonal directions in the $\mathbf{w}_c$ vector group. {PCA is performed only on the recorded critic weight trajectory collected at the end of each episode. No additional candidate or grid-sampled weights are included in the PCA computation.}

Now, we have two main orthogonal directions chosen to represent the $\mathbf{w}_c$ vector, which is $\boldsymbol{\delta}$ and $\boldsymbol{\eta}$ in the equation. To generate the full loss landscape, rather than a one-dimensional loss curve, the parameters $\alpha$ and $\beta$ are sampled over a grid. To calculate the loss corresponding to each combination of $\alpha$ and $\beta$ in the coordinate system, training data should be used for the calculation. In a supervised machine learning scenario, the training data is a batch of static data. However, in online reinforcement learning, the training data are collected and updated over time, \textbf{blue}{and thus differ across episodes. Therefore, to construct a well-defined loss landscape as a scalar field over the critic parameter space, the input data and corresponding targets must be fixed when scanning the parameter grid. } 

{Here, we consider that the data collected at each episode of training is one batch of data. The batch data in one episode is generated by the corresponding policy of that episode. In this study, unless otherwise specified, the reference batch dataset is selected as the states collected at each simulation time step of the final episode, and the corresponding targets are the temporal difference targets computed under the final policy.} With the weight value of each weight grid, the input states data and the target value, the loss corresponding to each weight grid is calculated. Hence, we derive a 3-D critic loss landscape which can depict the loss geometry around the final policy. By projecting the 3-D loss landscape to a 2-D contour with the same PCA plane, the recorded weight parameters formulate the optimization path of critic weights during training, which qualitatively shows the evolution of training relative to the landscape. {From the description of the method, principally, data from any episode could be choosen to visualize the loss landscape generated by the policy generated from the corresponding episode.} 

This construction yields a well-defined scalar field over the critic parameters with the inputs and targets fixed when scanning the grid. It therefore provides an interpretable view of the local geometry under the final policy and a qualitative depiction of the training trend through the overlaid path. It does not reproduce the exact per-episode temporal difference objective, which changes with the data and targets. But it serves as a performance index that reveals the geometry of the training landscape, which is relevant for explaining convergence or instability. {Other reference batch datasets, such as those from intermediate training stages, can also be used to generate landscape snapshots, as illustrated in \autoref{subsection: loss landscape during training}.}

\subsection{Quantitative analysis of loss landscape}
\label{subsection: loss landscape quantity analysis}
The critic match loss landscape is derived as in \autoref{subsection: ADHDP critic match loss method} and gives a qualitative view of the optimization path. To demonstrate the method beyond visual inspection and enable objective comparison, we introduce three quantitative indices.

\emph{sharpness}, \emph{basin area}, and \emph{local anisotropy} are introduced to depict the loss landscape quantitatively. They capture complementary aspects of the loss geometry around the final critic parameters.

Since loss scales differ across algorithms and systems, raw values are not directly comparable. Let $L(\alpha,\beta)$ be the loss on the PCA plane and $(\alpha^{*},\beta^{*})$ the final parameters with $L^{*}$. We define the relative surface
\begin{equation}
\Delta L(\alpha,\beta)=L(\alpha,\beta)-L^{*},
\end{equation}
and normalize it by the interquartile range (IQR) of $\Delta L$ over the grid to obtain a dimensionless surface $\tilde L$.

\paragraph{Sharpness}
Sharpness measures how fast the loss rises when moving away from the final point. For radius $\epsilon$,
\begin{equation}
\mathrm{Sharp}_{\epsilon}=
\max_{\theta\in[0,2\pi)}
\tilde L\!\left(
\alpha^{*}+\epsilon\cos\theta,\;
\beta^{*}+\epsilon\sin\theta
\right).
\end{equation}
A larger $\mathrm{Sharp}_{\epsilon}$ value indicates higher local stiffness that descent can be fast with small steps, while the stable step-size margin narrows and sensitivity to noise increases. A smaller $\mathrm{Sharp}_{\epsilon}$ value indicates a flatter, more tolerant neighborhood.

\paragraph{Basin area}
The basin area quantifies the extent of the low-loss set around the final point using
\begin{equation}
A_{\rho}=\mathrm{Area}\Big\{(\alpha,\beta)\ \big|\ \tilde L(\alpha,\beta)\le \rho\Big\}.
\end{equation}
A larger $A_{\rho}$ implies wider robustness around the final point while a very small or non-closed set indicates a fragile or non-convergent situation.

\paragraph{Local anisotropy}
Local anisotropy captures directional imbalance near $(\alpha^{*},\beta^{*})$. A quadratic fit of $\tilde L$ in a small neighborhood yields a $2\times2$ Hessian $H$. We use the condition number
\begin{equation}
\log\kappa=\log\!\left(\frac{\lambda_{\max}(H)}{\lambda_{\min}(H)}\right),
\end{equation}
where $\lambda_{\max}$ and $\lambda_{\min}$ are the largest and smallest eigenvalues of $H$. Large $\log\kappa$ indicates a narrow, ill-conditioned valley. It means that around the final point, one direction is steep while the other is flat, which makes progress sensitive to step size and update direction. Amall $\log\kappa$ corresponds to more isotropic curvature and smoother updates.

In conclusion, sharpness shows local stiffness and basin area describes tolerance. These two indices address different questions and should be interpreted with caution. Sharp landscapes can offer strong local attraction yet require tight step-size control, while wide basins indicate robustness to parameter perturbations. Anisotropy complements both by indicating how “skewed” the valley is. Together, these indices provide a compact quantitative description of the landscape. Combined with overlaid 2D-optimization path of the weight, they support comparisons across runs and help explain training behavior.

\subsection{System performance index}
\label{subsection:system performance index}
The critic match loss landscape will be analyzed quantitatively using the three indices in \autoref{subsection: loss landscape quantity analysis}. To relate the landscape geometry to the actual control performance of a dynamic system, a scalar system performance index is introduced.

To enable consistent comparison across different systems, each policy is evaluated on a fixed horizon $H$ using the same stage cost $c(x_t,u_t)$ as in training, which corresponds to the reinforcement signal ${r(t)}$ as in \autoref{eq:discounted cost function}. For performance evaluation, this instantaneous cost is normalized by a system-dependent upper bound, such that  $c(x_t,u_t)\in[0,1]$. Then we have
\begin{equation}
J_H=\sum_{t=0}^{H-1}\gamma^{t}\,c(x_t,u_t).
\end{equation}

If a failure occurs, such as states exceeding predefined constraints at time $t_{\mathrm{fail}}$, the rollout is continued to the horizon $H$ with a fixed worst-case penalty
\begin{equation}
c(x_t,u_t)=c_{\max},\qquad t\ge t_{\mathrm{fail}},
\end{equation}
where $c_{\max}=1$ corresponds to the maximum normalized stage cost.

Since $c(x_t,u_t)\in[0,1]$ is by construction, the discounted cost admits a natural upper bound. The performance index is therefore normalized as
\begin{equation}
\tilde J_H=\frac{J_H}{\sum_{t=0}^{H-1}\gamma^t}\in[0,1].
\end{equation}

From this definition, a well-controlled behavior yields a small $\tilde J_H$, whereas early failure results in a large $\tilde J_H$ due to the accumulated penalty. This single index enables direct comparison of system performance across tasks and remains meaningful even when training diverges, providing a unified basis for relating control performance
to the quantitative properties of the loss landscape.

\section{ADHDP control result for system with uncertainties}
\label{section: ADHDP controlr result}

In this section, the cart-pole system and the spacecraft attitude system are introduced. The control results of using the ADHDP algorithm are shown. The results will be used as examples for using the critic match loss landscape to interpret the online RL algorithm. {Multiple independent simulation runs were conducted for each system. For visualization and analysis, representative runs were selected to illustrate typical convergent and divergent learning behaviors.}

\subsection{Dynamics and control for cart-pole system}
\label{subsection: cart-poled dynamics and control}
\begin{figure}[H]
\centering
\includegraphics[width = 0.2\textwidth]{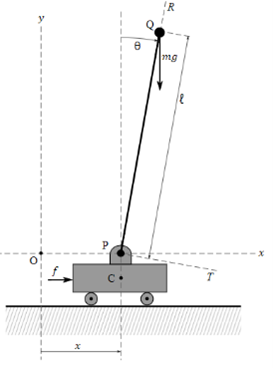}
\caption{Cart-pole system as a benchmark problem for ADHDP}
\label{fig:Cart-pole system}
\end{figure}

Consider the physical model shown in ~\autoref{fig:Cart-pole system}. A cart is positioned on a track running horizontally and an inverted pendulum is attached to the cart with a hinge that allows rotation around pivot point $P$ in the $x-y$ plane only, i.e. the pendulum is free to move horizontally. The pendulum's initial position is vertical, and an external variable horizontal force is applied to the cart to maintain the pendulum in a balanced and upright position. 

The model is considered to represent an unstable system. An external control force is required to keep the pendulum balanced, as opposed to a downward-hanging pendulum in which the force of gravity results in a stable oscillation about the downward vertical. An ideal pendulum is assumed in which all of the pendulum's mass is located at a single point ($Q$) at the end of a massless rigid rod.

The control goal of the cart-pole system is that by “pushing the cart to the right” or “pushing the cart to the left,” the pendulum can stay in its upright position. Since the force is added by pushing the cart, the input to the system is discrete with a constant value. 

The dynamics and kinematics of the system are described with the following equations~\citep{barto1983neuronlike}.
\begin{eqnarray}
\ddot{\theta}=\frac{g \sin \psi+\cos \psi\left[\frac{-f-m \hat{l} \dot{\psi}^2 \sin \psi+\mu_c \operatorname{sgn}(\dot{x})}{m_c+m}\right]-\frac{\mu_p \hat{\psi}}{m \hat{l}}}{\hat{l}\left\lfloor\frac{4}{3}-\frac{m \cos ^2 \psi}{m_c+m}\right\rfloor}
\label{eq:cpdynamics_theta}
\end{eqnarray}
% \begin{eqnarray}
% \ddot{x}=\frac{f+m \hat{l}\left[\dot{\theta}^2 \sin \theta-\bar{\theta} \cos \theta\right]-\mu_c \operatorname{sgn}(\dot{x})}{m_c+m}
% \label{eq:cpdynamics_x}
% \end{eqnarray}
\begin{eqnarray}
\begin{aligned}
& \ddot{x}=\frac{f+m \hat{l}\left[\dot{\psi}^2 \sin \psi-\bar{\psi} \cos \psi\right]-\mu_c \operatorname{sgn}(\dot{x})}{m_c+m} \\
& g = -9.8\,\mathrm{m/s^2}
\end{aligned}
\label{eq:cpdynamics_x}
\end{eqnarray}
where $\psi$ is the angular placement of the pendulum, and $x$ is the horizontal placement of the cart, as indicated in ~\autoref{fig:Cart-pole system}. $f$ is the force that is applied to the cart, $m$ is the mass of the pendulum, $m_c$ is the mass of the cart, $\mu_p$ and $\mu_c$ are the friction coefficients for the pendulum and the cart, and $\hat{l}$ is the length of the pendulum.

The ADHDP algorithm is first applied to the cart-pole dynamic system. Then, the critic loss landscape visualization method is applied to reveal the training process of the algorithm.

For the ADHDP control of the cart-pole system, the state vector is \( [\psi, \dot{\psi}, x, \dot{x}] \), and the control input is \( [f] \). The input to the critic neural network is the combination of the state vector and the control input. The input to the actor neural network is the state vector. The MLP structure with one hidden layer is chosen to approximate the critic and actor neural networks. The number of neurons of the hidden net is 6 and 4, respectively. {The actor and critic learning rates are set to $1 \times 10^{-3}$ and $1 \times 10^{-2}$ respectively. Training is conducted over 100 trials, or episodes.  The ADHDP algorithm is trained in an online manner, where at each dynamic simulation time step the current state is sampled and used immediately to compute the temporal-difference error and update the actor and critic networks. No minibatch training is employed, which is equivalent to a batch size of one. Network weights are initialized using uniformly distributed random values. No explicit exploration noise is added to the actor output in the baseline ADHDP implementation as shown in \autoref{Appendix}, in order to avoid introducing additional stochasticity that could confound the interpretation of critic optimization behavior.}

\begin{figure}[H]
     \centering
     \begin{subfigure}[b]{0.3\textwidth}
         \centering
         \includegraphics[width=\textwidth]{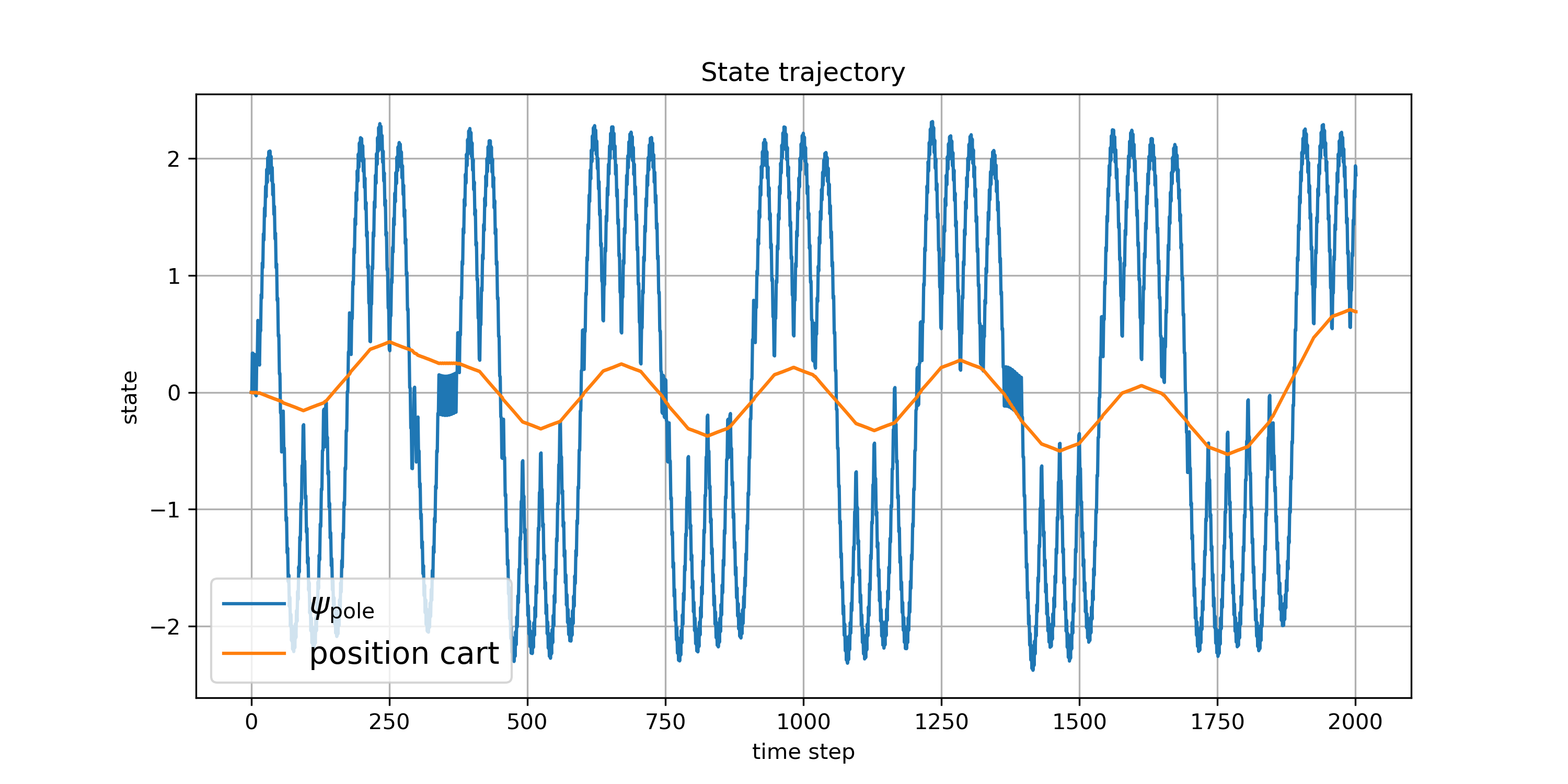}
         \caption{State trajectory}
         \label{fig:cp_iniweight_2nd_a}
     \end{subfigure}
     \hfill
     \begin{subfigure}[b]{0.3\textwidth}
         \centering
         \includegraphics[width=\textwidth]{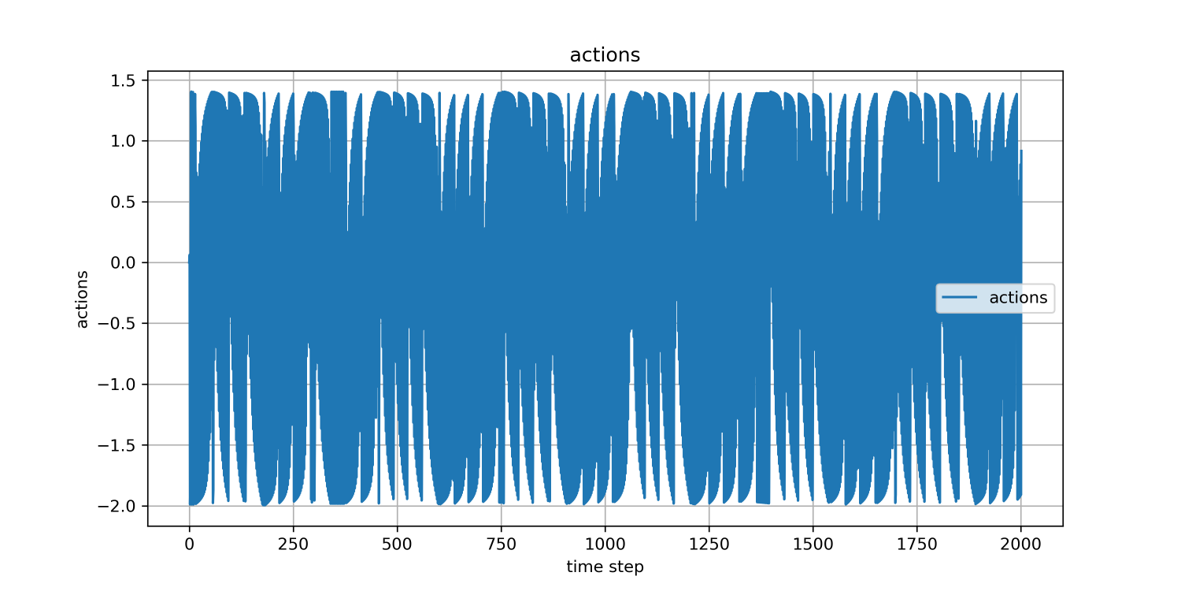}
         \caption{Actions}
         \label{fig:cp_iniweight_2nd_b}
     \end{subfigure}
     \hfill
     \begin{subfigure}[b]{0.3\textwidth}
         \centering
         \includegraphics[width=\textwidth]{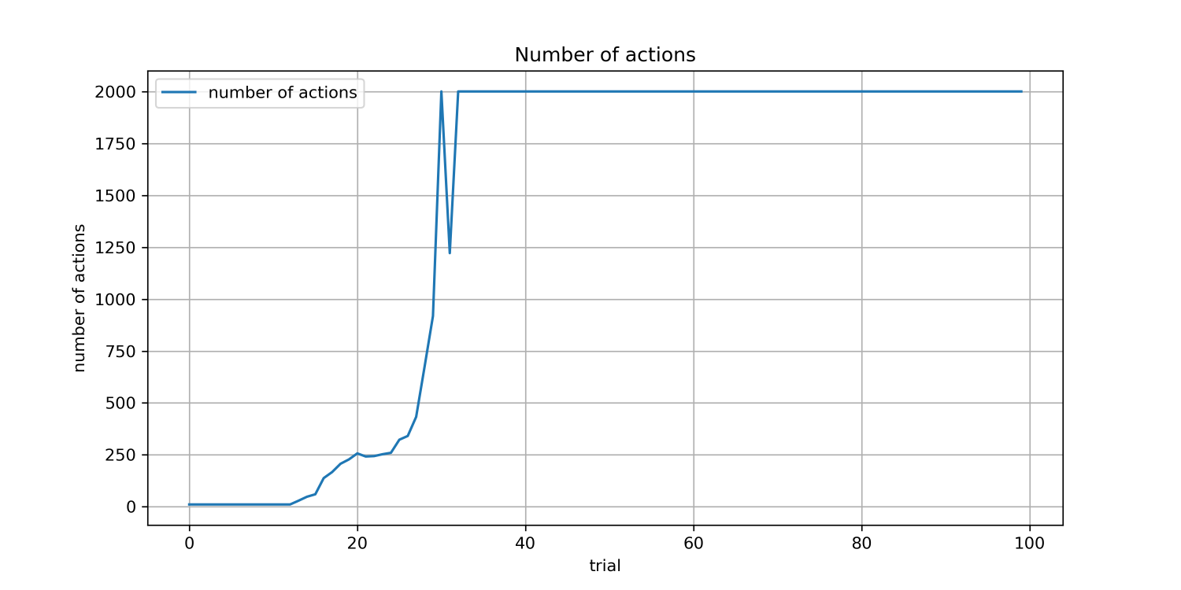}
         \caption{Number of trials}
         \label{fig:cp_iniweight_2nd_c}
     \end{subfigure}
        \caption{Cart-pole control results}
        \label{fig:cp_iniweight_2nd}
\end{figure}

\autoref{fig:cp_iniweight_2nd} shows a successful simulation result of using ADHDP to control the cart-pole system. \autoref{fig:cp_iniweight_2nd_c} shows how many actions the system can take during each trial before it diverges in 100 trials of running the simulation. \autoref{fig:cp_iniweight_2nd_a} shows the evolution of the states in the 100$^{th}$ trial, which can be maintained at the desired range. \autoref{fig:cp_iniweight_2nd_b} shows the corresponding actions the system takes.

\begin{figure}[H]
    \centering
    \begin{subfigure}[b]{0.45\textwidth}
        \centering
        \includegraphics[width=\textwidth]{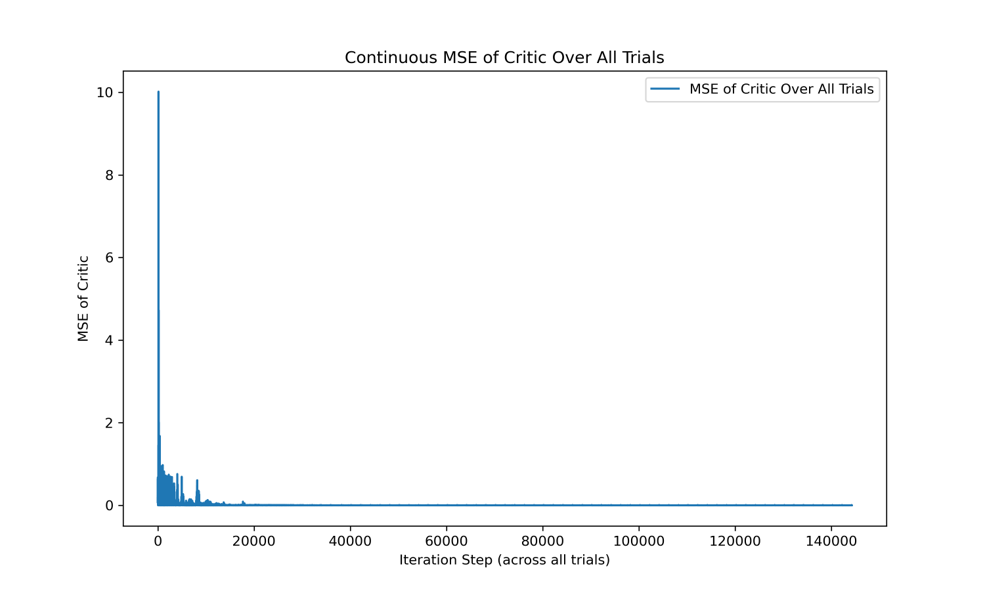}
        \caption{Critic loss}
        \label{fig:cp_critic loss}
    \end{subfigure}
    \hspace{0.02\textwidth}  % Small space between the two subfigures
    \begin{subfigure}[b]{0.45\textwidth}
        \centering
        \includegraphics[width=\textwidth]{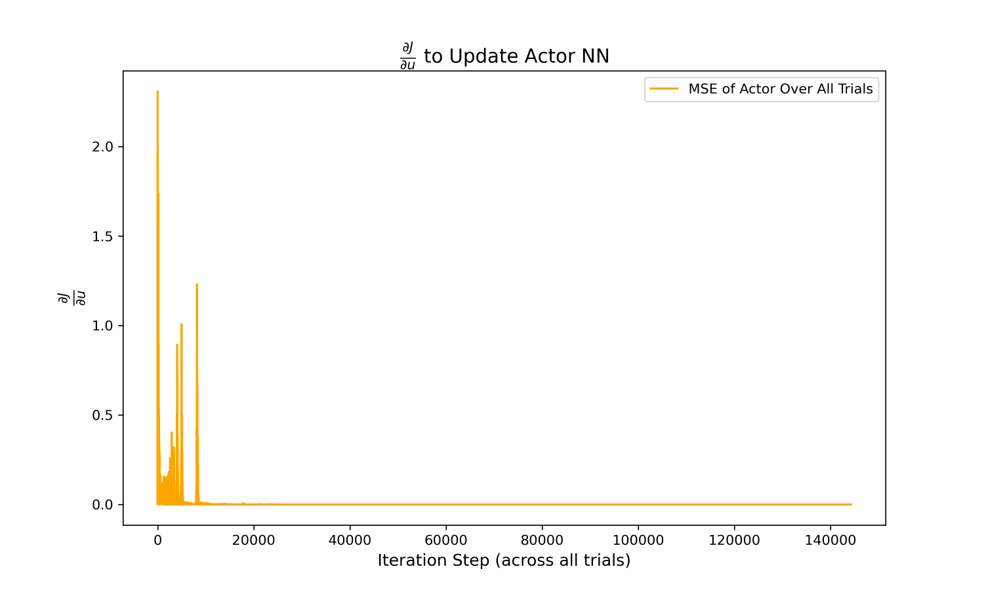}
        \caption{Actor gradient w.r.t. J}
        \label{fig:cp_actor gradient}
    \end{subfigure}
    \caption{Critic and actor NN evolution process with time, cart-pole system}
    \label{fig:cp_training process time}
\end{figure}

\autoref{fig:cp_critic loss} shows the critic match loss during training. It can be seen that, as the training continues, the critic match loss decreases to around zero. It means that the critic network is converging. \autoref{fig:cp_actor gradient} shows the evolution of $\frac{\partial \mathrm{J}(t)}{\partial \mathrm{u}(t)} $.This item is used to update the actor neural net, as shown in \autoref{eq:actor_BBPi_2}. As the training continues, the gradient decreases to around zero. It indicates it reaches a sub-optimal solution.

\subsection{Attitude dynamics and control for spacecraft with unknown inertia}

In section 3.1, the visualization method has been validated with the benchmark cart-pole system. In this part, the results of applying the algorithm to the spacecraft attitude control system, a more complex problem with unknown inertia, will be studied.

Following the benchmark cart-pole system, the combined spacecraft dynamic system will be introduced in this part. From the literature review in \autoref{sec1}, the combined spacecraft in the ADR scenario is a system with uncertainties. In order to progressively test the performance of the algorithm on the control of the post-capture scenario, the combined spacecraft model with the lowest level of uncertainty is considered in the work, which is the uncertainty of the unknown inertial parameter. Based on this, the following assumptions are made. The target is captured tightly by rigid robotic manipulators on the rigid service spacecraft. The joints of spacecraft manipulators will be locked once the target spacecraft is captured \citep{huang2018postcapture}. The target is rigid and uncooperative without control capability. Under the given assumptions, the combined spacecraft is taken as a single rigid spacecraft with unknown inertia parameters. 

The kinematic and dynamic equations of combined spacecraft in the inertia principal axis frame of combination use Euler angles. In this paper, a body-fixed reference frame $\mathcal{B}$ is defined, located at the center-of-mass of the combined spacecraft. Its coordinate axes are aligned along the principal directions of the spacecraft. In addition, an Earth-Centered Inertial (ECI) coordinate frame $\mathcal{N}$ is defined, with unit vectors $\left\{\bm{\overline{n_1}},\bm{\overline{n_2}}, \bm{\overline{n_3}}\right\}$.The $\bm{\overline{n_1}}$ axis is aligned with the mean equinox. The $\bm{\overline{n_3}}$ axis is aligned with the Earth's rotation axis or the celestial North Pole, and the $\bm{\overline{n_2}}$ complies with the right-handed system.

For the rotation from frame $\mathcal{N}$ to frame $\mathcal{B}$ with the sequence of 3-2-1, the kinematics of the combined spacecraft are given below. In the scenario for the combined spacecraft stabilization, the assumption is that the initial attitude states of the combined spacecraft are not expected to deviate significantly from zero degrees, with the final control goal as zero degrees. Hence, the combined spacecraft here is represented with attitude angles, which will not suffer from the singularity problem under the given scenario. 
\begin{eqnarray}
\left[\begin{array}{c}
\dot{\theta}_1 \\
\dot{\theta}_2 \\
\dot{\theta}_3
\end{array}\right]=\frac{1}{\cos \theta_2}\left[\begin{array}{ccc}
\cos \theta_2 & \sin \theta_1 \sin \theta_2 & \cos \theta_1 \sin \theta_2 \\
0 & \cos \theta_1 \cos \theta_2 & -\sin \theta_1 \cos \theta_2 \\
0 & \sin \theta_1 & \cos \theta_1
\end{array}\right]\left[\begin{array}{c}
\omega_1 \\
\omega_2 \\
\omega_3
\end{array}\right]
\label{eq:spc_kinematics}
\end{eqnarray}
The attitude dynamics is given as 
\begin{eqnarray}
\bar{M}=\hat{J}_{s c} \cdot \dot{\bar{\omega}}+\bar{\omega} \times \hat{J}_{s c} \cdot \bar{\omega}
\label{eq:spc_dynamics}
\end{eqnarray}
Here, \( \theta_1 \), \( \theta_2 \), and \( \theta_3 \) represent the three attitude angles of the spacecraft, and \( \omega_1 \), \( \omega_2 \), and \( \omega_3 \) are the corresponding angular velocities. \( \bar{M} \) denotes the control torque applied to the attitude system. \( \hat{J}_{\mathrm{sc}} \) is the inertia matrix of the spacecraft, which is assumed to be unknown.

The reward function is given in the form of a quadratic function of state errors and control input, as follows.
\begin{eqnarray}
r(x, u)=x^T P x+u^T Q u
\label{eq:reward function}
\end{eqnarray}
 where \( x = \left[ e_{\theta_1}, e_{\theta_2}, e_{\theta_3}, e_{\omega_1}, e_{\omega_2}, e_{\omega_3} \right] \) represents the error between the observed and desired states. For the attitude stabilization task, the desired states are all equal to zero. {During this simulation, value of $P$ is 0.01, value of Q is set to zero to reduce the complexity of the learning problem. The inertial parameter of the combined spacecraft is described with the matrix $\hat{J}_{s c}$ with value $[1 0.1 0.1; 0.1 0.1 0.1; 0.1   0.1 0.9] \si{kg.m^{2}}$.}

\begin{figure}[H]
     \centering
     \begin{subfigure}[b]{0.3\textwidth}
         \centering
         \includegraphics[width=\textwidth]{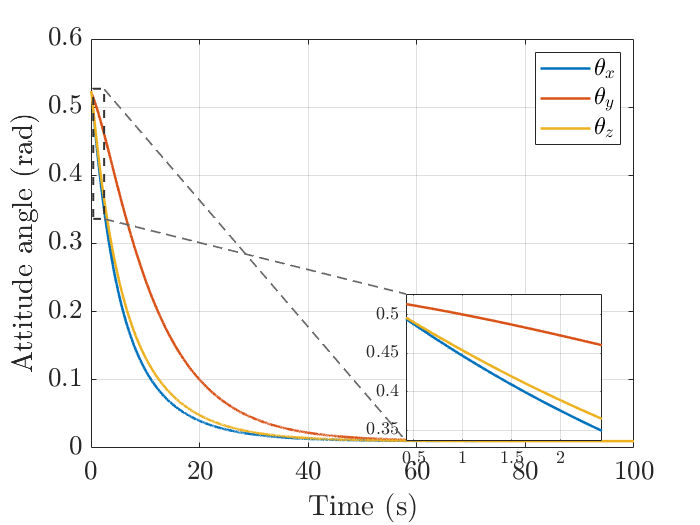}
         \caption{Attitude angle}
         \label{fig:spc_PD_a}
     \end{subfigure}
     \hfill
     \begin{subfigure}[b]{0.3\textwidth}
         \centering
         \includegraphics[width=\textwidth]{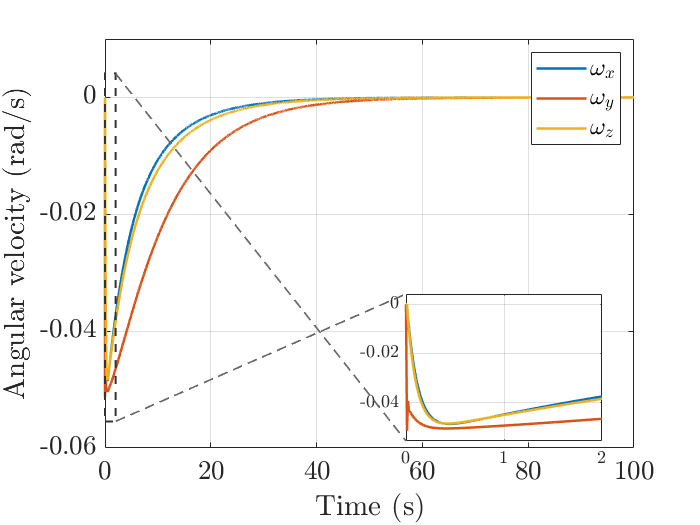}
         \caption{Angular rate}
         \label{fig:spc_PD_b}
     \end{subfigure}
     \hfill
     \begin{subfigure}[b]{0.3\textwidth}
         \centering
         \includegraphics[width=\textwidth]{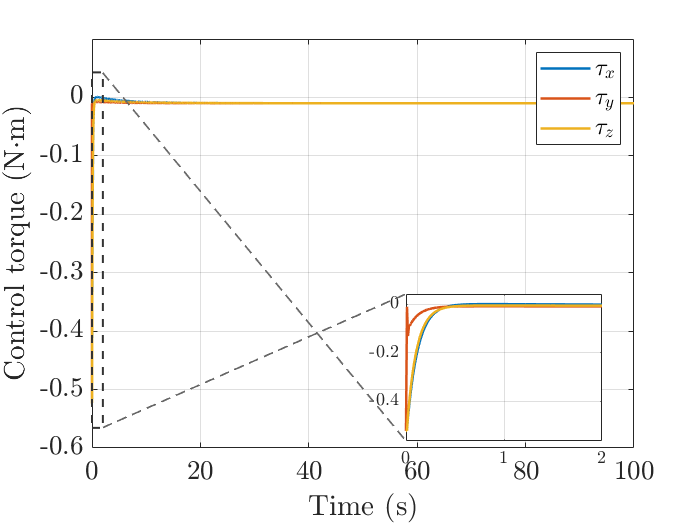}
         \caption{Control torque}
         \label{fig:spc_PD_c}
     \end{subfigure}
        \caption{Results with PD control}
        \label{fig:spc_PD}
\end{figure}

As a reference for the performance of the ADHDP, we start by applying standard PD control to the problem. The control gains $K_p$ and $K_d$ are selected as $\operatorname{diag}(1, 1, 1)$ and $\operatorname{diag}(10, 10, 10)$, respectively.
The results are shown in ~\autoref{fig:spc_PD}.

As can be seen from \autoref{fig:spc_PD}, the system can be stabilized in the given 100 seconds. And the control torque does not exceed 0.1 Nm. 

The ADHDP algorithm is applied to the above-defined spacecraft attitude dynamic system. Then the critic match loss landscape visualization method is applied to reveal the training process of the algorithm.

For the ADHDP control of the spacecraft attitude dynamic system, the state vector is $\left[\theta_1, \theta_2, \theta_3, \omega_1, \omega_2, \omega_3\right]$, and the control input is $u$. The input to the critic neural network is the combination of the state vector and the control input. The input to the actor neural network is the state vector. The MLP structure with one hidden layer is chosen to approximate the critic and actor neural networks. The number of neurons of the hidden net is 10 for both neural networks.

The hyperbolic tangent function is used as the activation function for the hidden layer in both actor and critic neural networks to add nonlinearity to the approximation process. For the selection of the activation function for the output layers of the neural network, the linear function is selected as the output function of the critic network, considering the range of the linear function and the cost function. The hyperbolic tangent function is used as the output activation of the actor network, since its range sufficiently accommodates the required control torque, which is less than 1 Nm as shown in \autoref{fig:spc_PD_c}. The actor and critic neural networks are initialized using the default initialization of the linear layers. 

{The actor and critic learning rates are set to $1 \times 10^{-3}$ and $1 \times 10^{-2}$ respectively. The critic learning weight is adaptively adjusted during training using a multiplicative scaling ratio of 1.01, with an upper bound of 1.2. The algorithm is trained in an online manner following the same update scheme as the cart-pole case in \autoref{subsection: cart-poled dynamics and control}. Parameter updates are performed at each simulation time step, corresponding to a batch size of one, and no explicit exploration noise is added.}

{Training is conducted over 300 episodes. For each episode, using the PD control results in \autoref{fig:spc_PD} as a reference, the boundary for simulation time step is set for 100 seconds.} For each time step in the episode, the actor and critic networks are trained in 30 epochs. To efficiently train the algorithm, a limit is set to the states of the system, which means that once the system starts to diverge and the states exceed the limit, the current episode will terminate. The limit can be set according to the simulation scenario. For example, a certain value can be given to the angular rate as the limit. {For diverging runs, the final episode used for loss landscape construction corresponds to the last executed episode before training termination, which may be an early-terminated episode due to state divergence.}
\begin{figure}[H]
    \centering
    \begin{subfigure}[t]{0.32\textwidth}
        \centering
        \includegraphics[width=\textwidth]{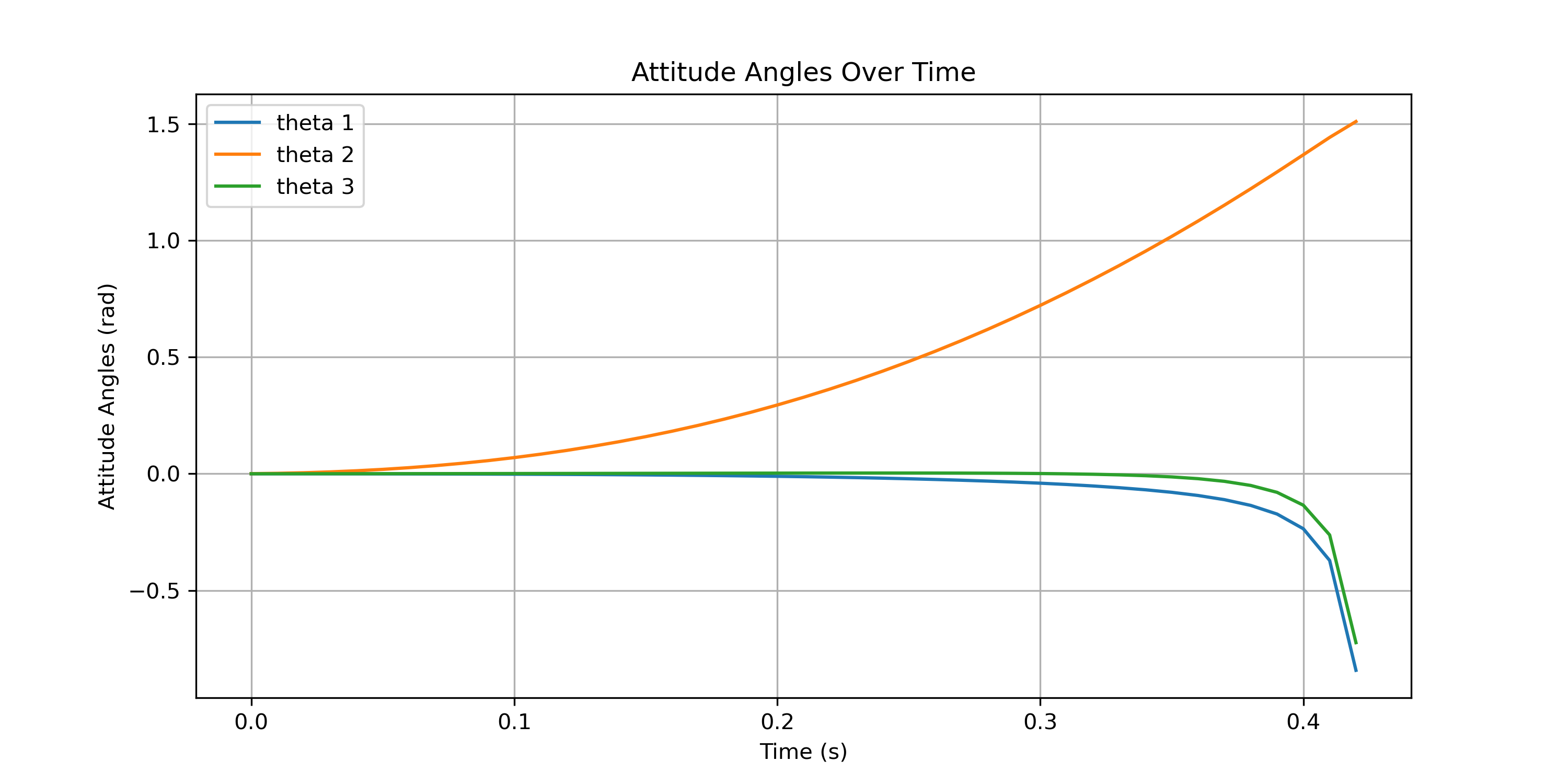}
        \caption{Attitude angle}
        \label{fig:spc_ADHDP_la_a}
    \end{subfigure}
    \hspace{0.005\textwidth}
    \begin{subfigure}[t]{0.32\textwidth}
        \centering
        \includegraphics[width=\textwidth]{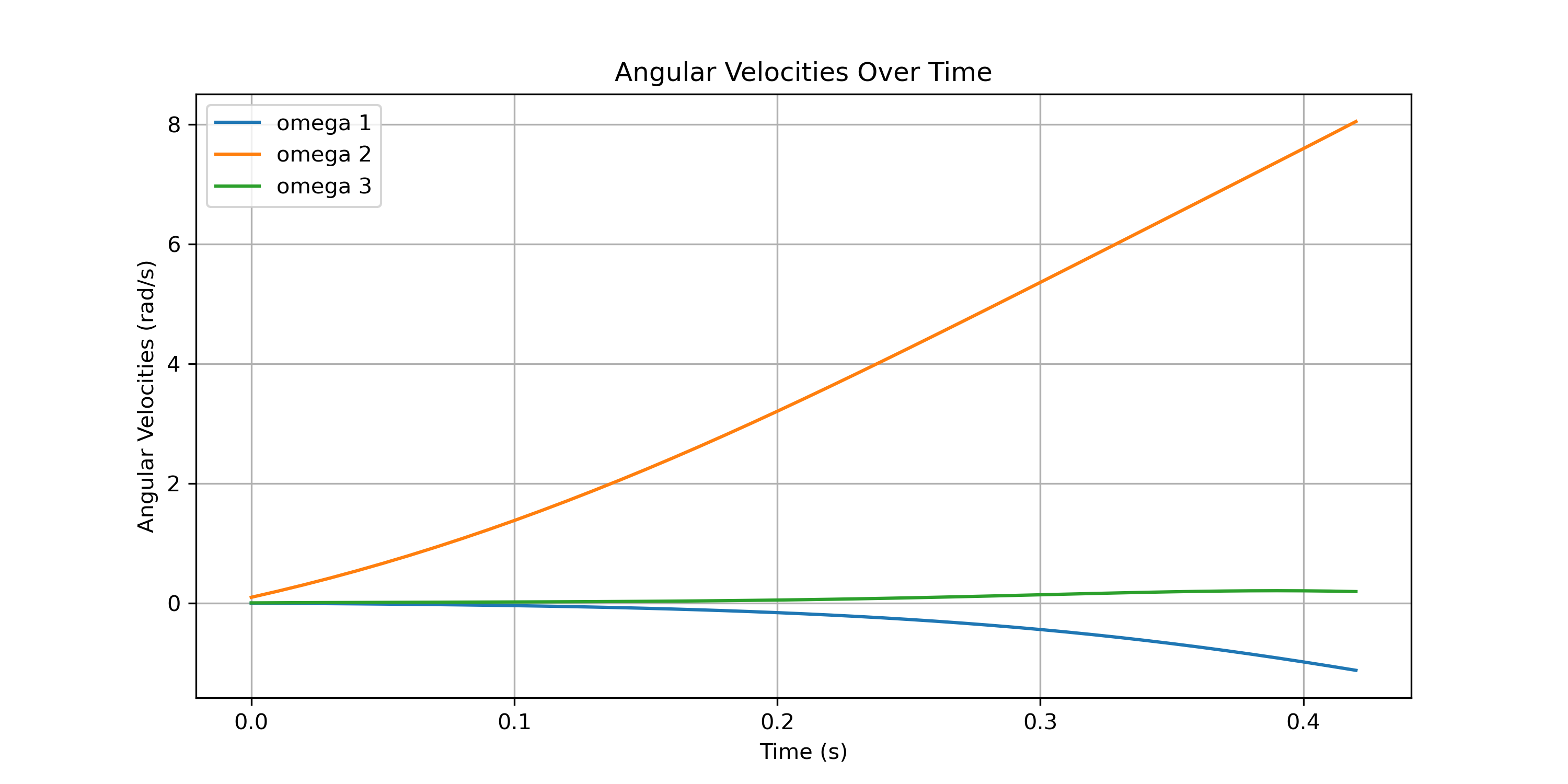}
        \caption{Angular rate}
        \label{fig:spc_ADHDP_la_b}
    \end{subfigure}
    \hspace{0.005\textwidth}
    \begin{subfigure}[t]{0.32\textwidth}
        \centering
        \includegraphics[width=\textwidth]{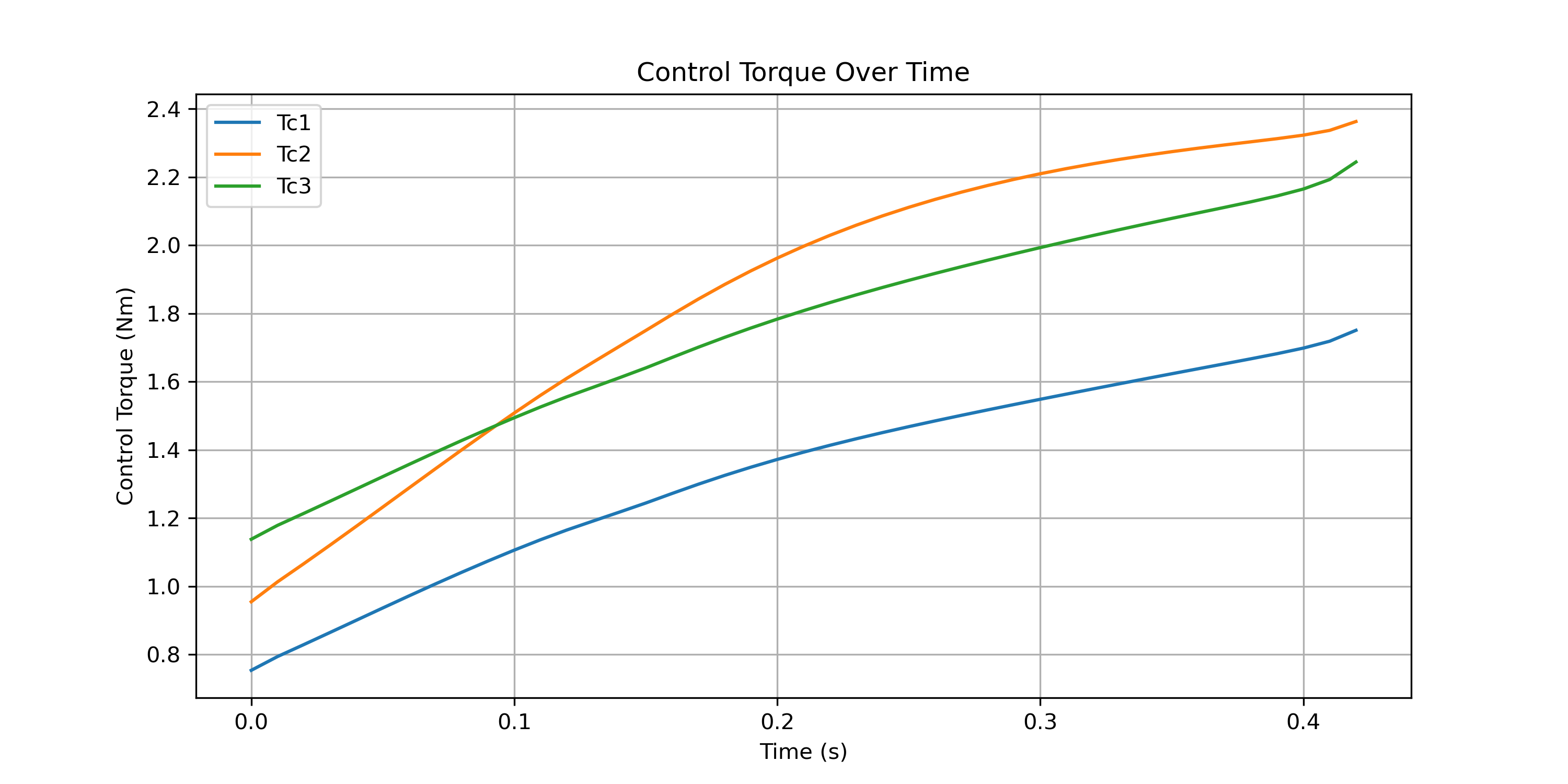}
        \caption{Control torque}
        \label{fig:spc_ADHDP_la_c}
    \end{subfigure}
    \caption{Spacecraft control results with ADHDP}
    \label{fig:spc_ADHDP_la}
\end{figure}

\begin{figure}[H]
    \centering
    \begin{subfigure}[b]{0.45\textwidth}
        \centering
        \includegraphics[width=\textwidth]{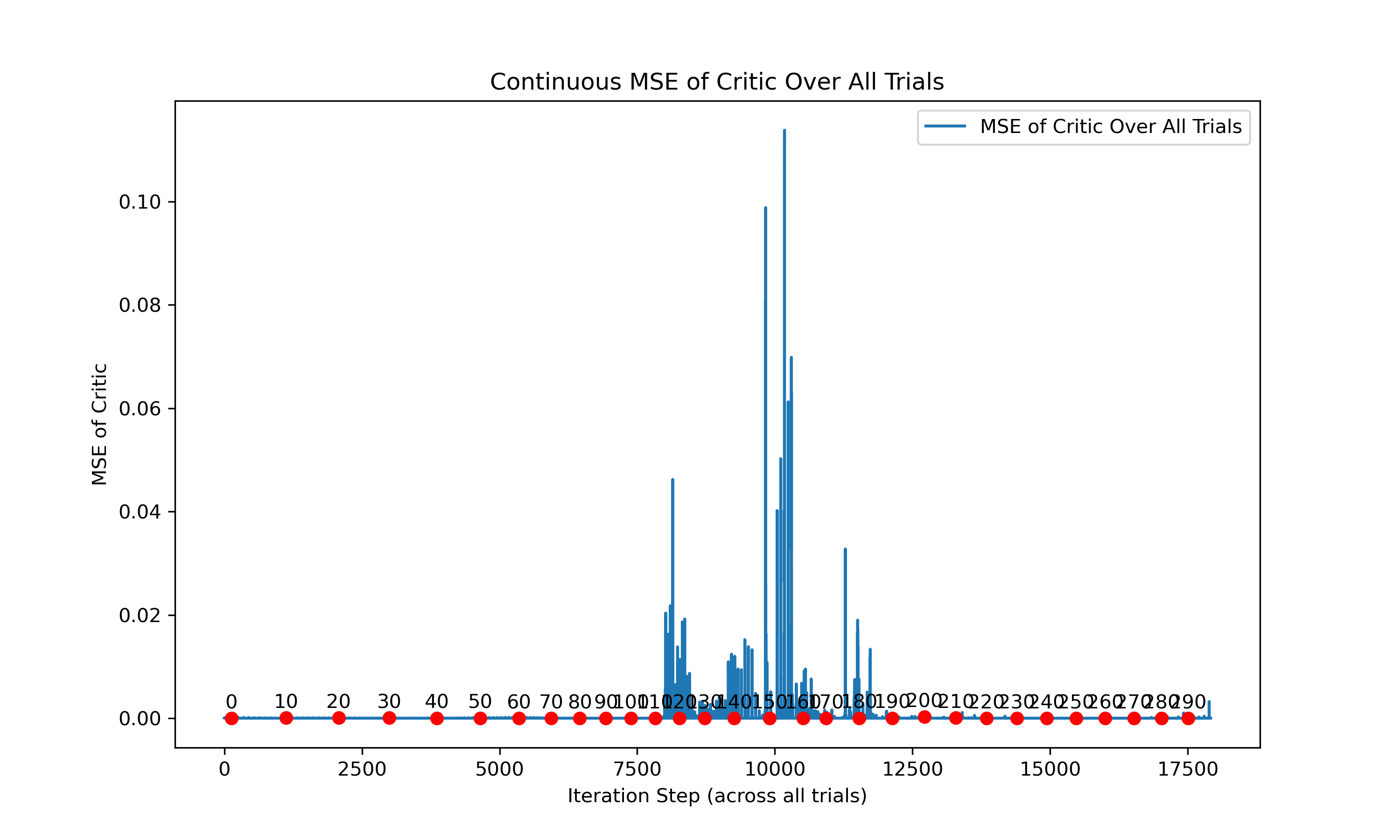}
        \caption{Critic loss}
        \label{fig:spc_critic loss}
    \end{subfigure}
    \hspace{0.02\textwidth}  % Small space between the two subfigures
    \begin{subfigure}[b]{0.45\textwidth}
        \centering
        \includegraphics[width=\textwidth]{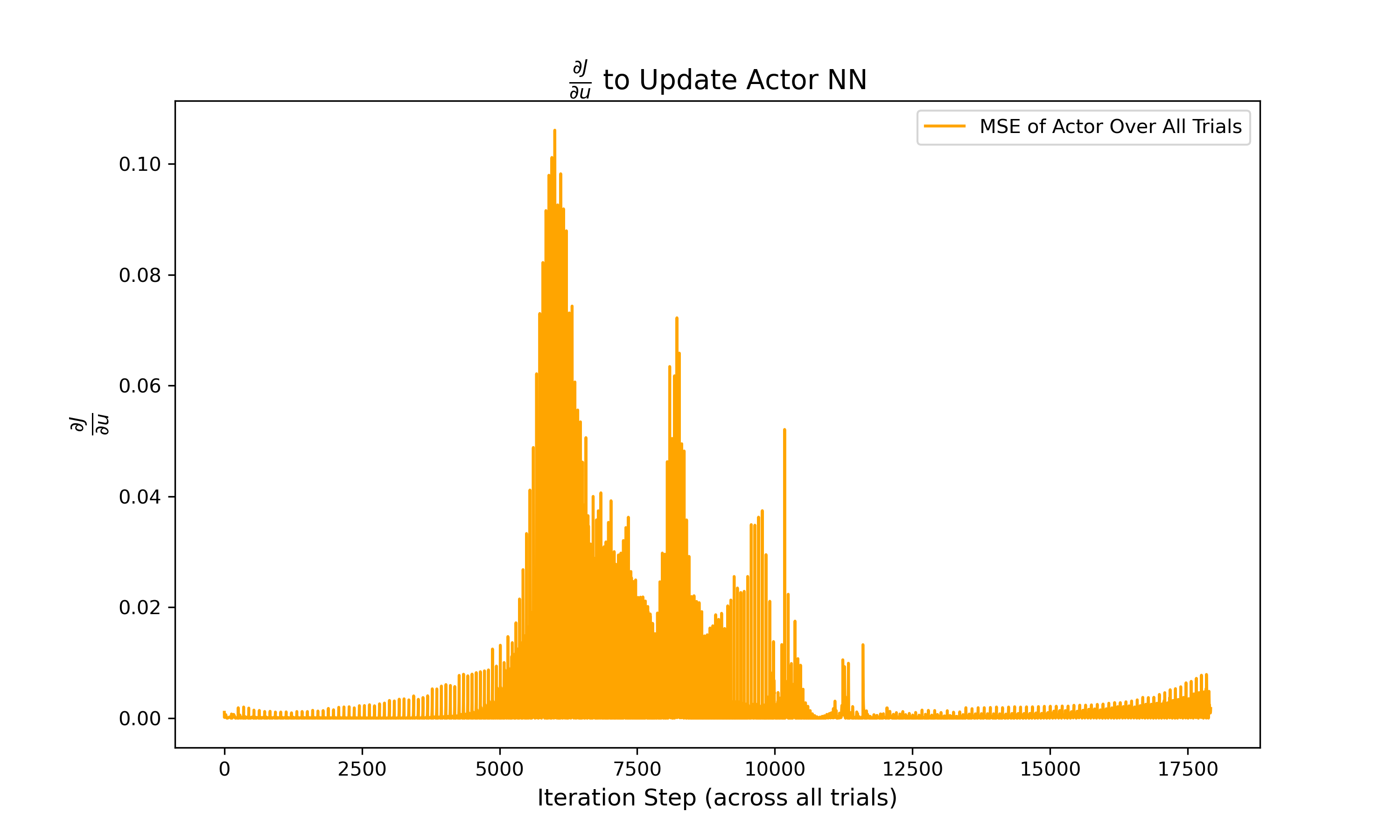}
        \caption{Actor gradient w.r.t. J}
        \label{fig:spc_actor gradient}
    \end{subfigure}
    \caption{Critic and actor NN evolution process with time, spacecraft system}
    \label{fig:spc_training process time}
\end{figure}

\autoref{fig:spc_ADHDP_la} is the control result of applying the ADHDP algorithm to the spacecraft attitude system. Apparently, the algorithm doesn't manage to successfully control the system. 

\autoref{fig:spc_training process time} shows the critic match loss and the evolution of $\frac{\partial \mathrm{J}(\mathrm{t})}{\partial \mathrm{u}(\mathrm{t})} $ during the training process of the ADHDP algorithm in the spacecraft attitude control scenario. The critic loss remains low during the early phase of training, followed by a series of sharp spikes in the middle phase. In the final phase, the loss stabilizes again at a relatively low level, significantly below the peak values observed earlier.The actor gradient $\frac{\partial \mathrm{J}(\mathrm{t})}{\partial \mathrm{u}(\mathrm{t})}$ uses critic loss for calculation. So a similar trend is seen in \autoref{fig:spc_actor gradient}: a gradual increase leading to several pronounced peaks during the middle phase, followed by smaller and more stable gradient values toward the end.  However, when the training reaches the stop limit, the actor loss is still high, with an increasing trend, which is definitely indicating divergence.

\section{Critic match loss landscape visualization for ADHDP algorithm}

In this section, the results of using the visualization method on the ADHDP algorithms are shown. The performance of the algorithms will be integrated using the critic match loss landscape.

\subsection{Critic match loss landscape visualization for cart-pole ADHDP control {under final policy}}

\begin{figure}[H]
    \centering
    \begin{subfigure}[b]{0.45\textwidth}
        \centering
        \includegraphics[width=\textwidth]{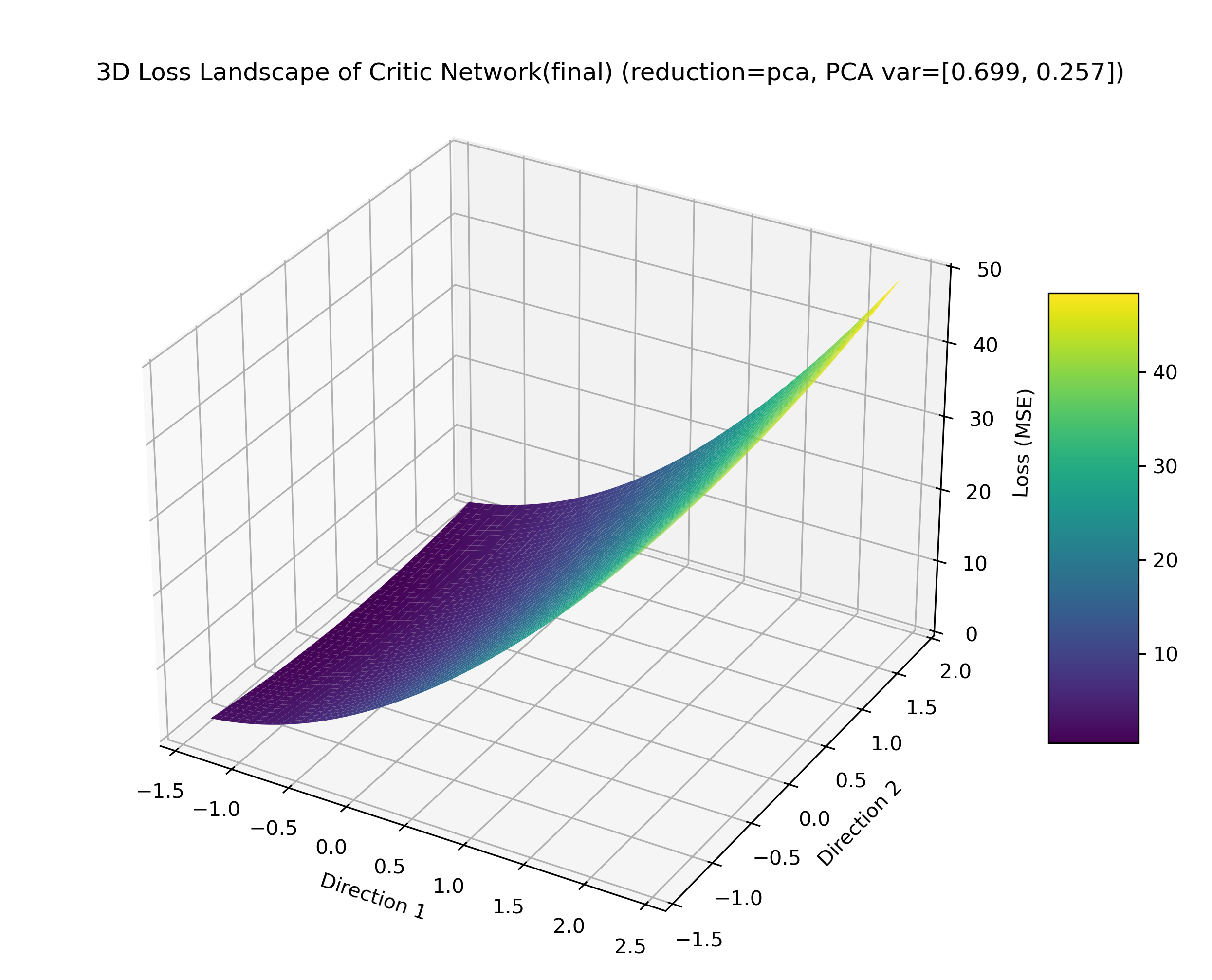}
        \caption{3-D loss of cart-pole ADHDP control  under final policy }
        \label{fig:cp_3D loss}
    \end{subfigure}
    \hspace{0.02\textwidth}
    \begin{subfigure}[b]{0.45\textwidth}
        \centering
        \includegraphics[width=\textwidth]{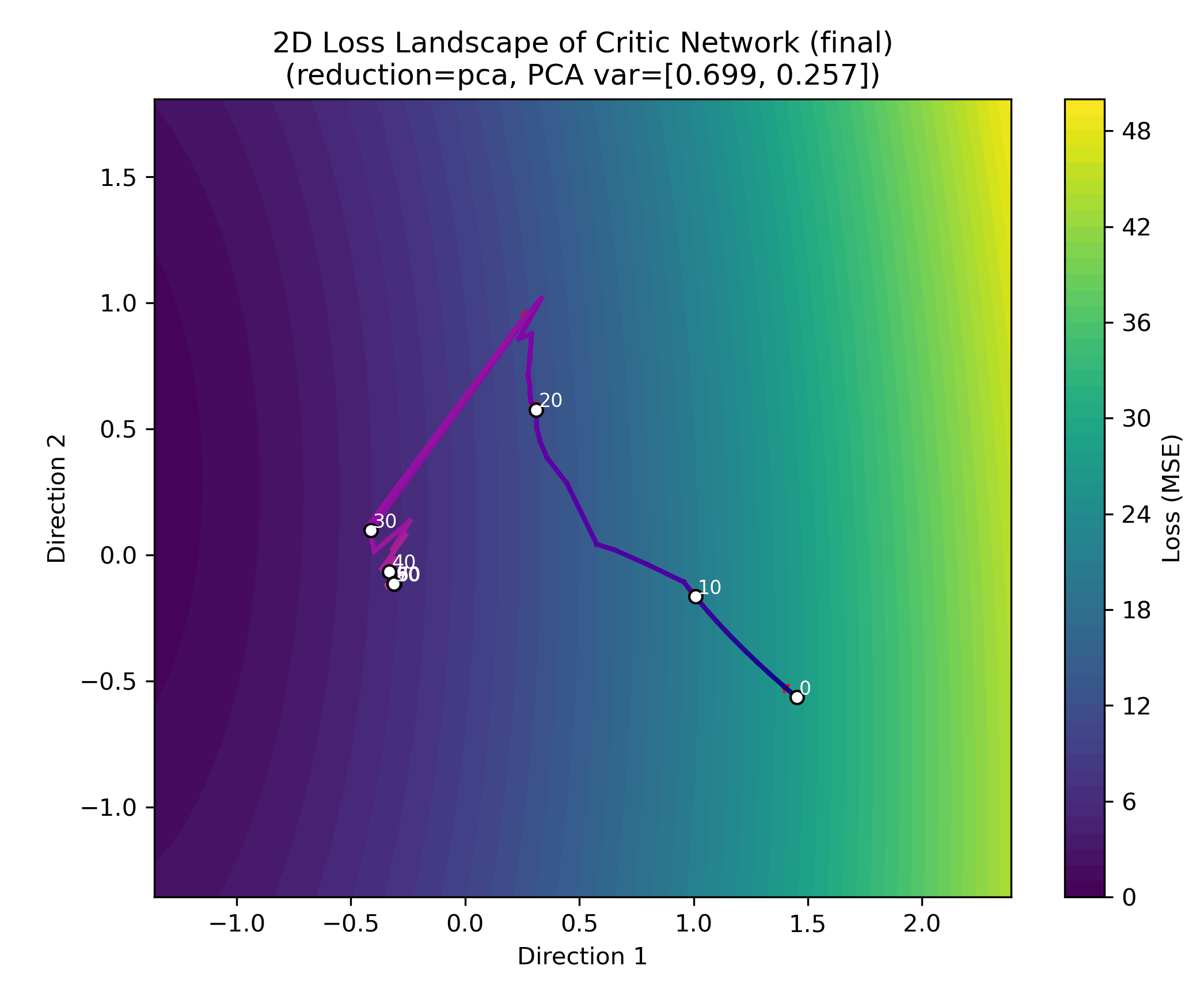}
        \caption{2-D loss path of cart-pole ADHDP control under final policy}
        \label{fig:cp_2D_loss}
    \end{subfigure}
    \caption{3-D and 2-D loss landscape of cart-pole ADHDP control {under final policy} }
    \label{fig:cp_loss landscape}
\end{figure}

\autoref{fig:cp_3D loss} shows how the critic loss landscape changes with weight of the critic NN. According to section 3.1, the critic network is a MLP with one hidden layer. The combined weight matrix is a  vector with multi-dimensions . After PCA is applied, the dimension of the matrix is recuded to two dimensions, which are illustrated by the x and y axes. The z axis indicates the loss across the weight grid on the x and y axes. The figure then shows the loss trend under different value combinations of the weight vector. Since the usage of the weight is to approximate the loss function,  the 3-D loss landscape can be used to analyze the characteristics of the algorithm revolution. From the example of using ADHDP to control the cart-pole system, it can be seen that the training brings a smooth loss landscape. Under this smooth loss landscape, the training process produces an accurate approximation of the loss function. This approximation of the cost function supports successive updates of the weight parameters. {For the convergent cart-pole case shown in \autoref{fig:cp_3D loss}, the first two principal components explain 69.9\% and 25.7\% of the variance of the critic weight trajectory, respectively. Together, the first two principal components capture 95.6\% of the total variance, indicating that the dominant directions of critic parameter evolution are well represented in the two-dimensional visualization plane.}

\autoref{fig:cp_2D_loss} is a 2-D projection of the 3-D loss landscape. The line in the projection indicates the update path of the weight during training. The beginning of the weight updates from a value that results in a relatively large critic match loss. As the training continues, the loss is optimized following the gradient descent of the weights. Finally, the weight update converges to a value that results in a low critic match loss. To visualize the update trajectory, dots are plotted along the 2-D path at regular intervals (e.g., every 10 time steps). The time step distance between every neighbor dot is equal. However, the change of critic match loss between every neighboring dot is different. It reveals the trend that the critic loss changes fast in the beginning, slows down in the middle, and hardly changes in the final. This pattern is consistent with gradient-based optimization, where large initial gradients drive fast loss reduction, followed by smaller updates near convergence. When the update stops, the process doesn't necessarily reach the global optimum, but it settles at a point where the critic loss is low enough to give good control performance. We call this a sub-optimal point, which is common in online learning where exact optimization isn't always possible. This pattern corresponds with \autoref{fig:cp_critic loss}, which depicts the critic loss changes with iteration steps. The critic loss changes steeply in the early steps. After the sudden decrease, the critic loss stays around the convergent value, which indicates the weight behaviour around the sub-optimal point.

\subsection{Critic match loss landscape visualization for spacecraft attitude control {under final policy}}

\begin{figure}[H]
    \centering
    \begin{subfigure}[b]{0.45\textwidth}
        \centering
        \includegraphics[width=\textwidth]{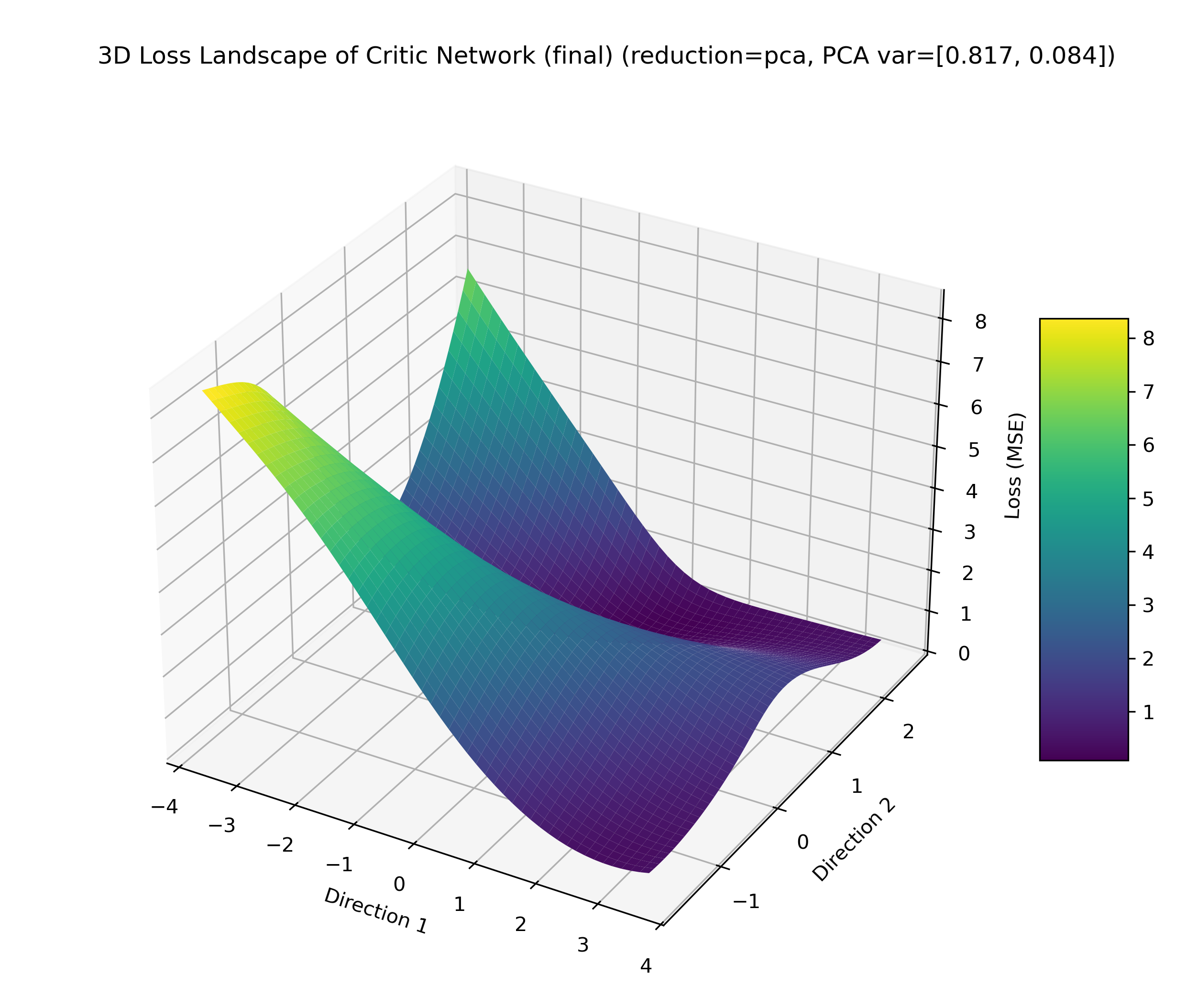}
        \caption{3-D loss landscape}
        \label{fig:spc_3D loss}
    \end{subfigure}
    \hspace{0.02\textwidth}
    \begin{subfigure}[b]{0.45\textwidth}
        \centering
        \includegraphics[width=\textwidth]{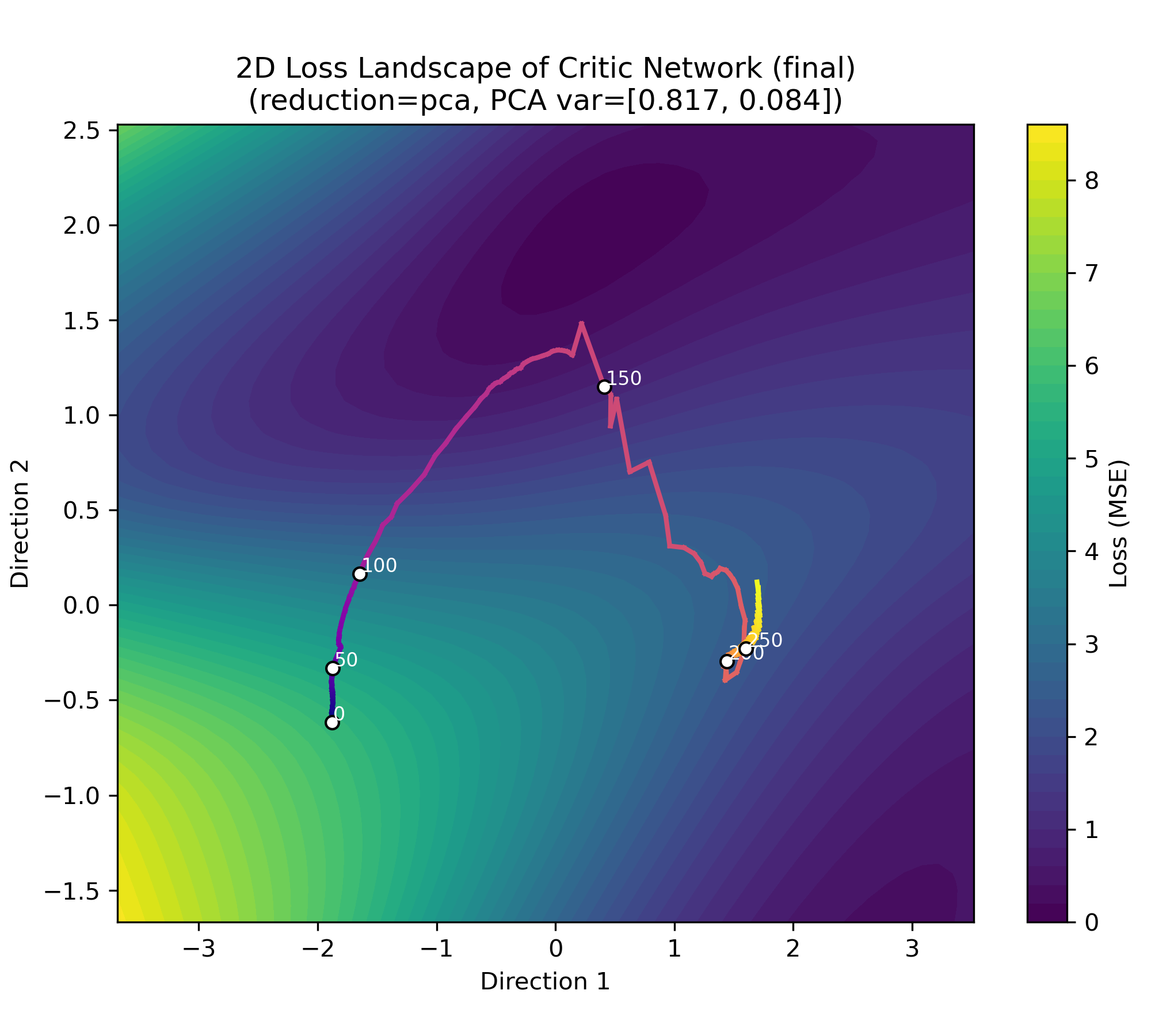}
        \caption{2-D loss landscape}
        \label{fig:spc_2D_loss}
    \end{subfigure}
    \caption{3-D and 2-D loss landscape of spacecraft attitude ADHDP control {under final policy}}
    \label{fig:spc_loss landscape}
\end{figure}

\autoref{fig:spc_loss landscape} illustrates the 3-D critic match loss landscape and the 2-D loss path resulting from applying the ADHDP algorithm to the spacecraft attitude system. {Note that for this diverging case, the loss landscape is constructed using the final executed episode, which terminates early due to state divergence. From \autoref{fig:spc_3D loss}, it can be seen that the loss landscape based on the final policy yields a terrain which has two peak-like structure and two bowl-shaped regions. The optimization path of the critic weight travels from peak to one of the bowls and ends in the other bowl, which corresponds to the fluctuations in loss values in \autoref{fig:spc_critic loss}.  For the divergent spacecraft attitude control case shown in \autoref{fig:spc_3D loss}, the first two principal components explain 81.7\% and 8.4\% of the variance of the critic weight trajectory, capturing 90.1\% of the total variance. The PCA results further indicate that the critic parameter updates exhibit strong anisotropy, with the first principal component accounting for 81.7\% of the total variance. This implies that the evolution of the critic parameters during training is dominated by updates along a single principal direction in the parameter space. If this dominant update direction does not correspond to a descent direction of the true cost function, which may arise from unstable or rapidly varying TD target, the accumulated updates may drive the learning process toward divergence rather than convergence. This behavior is consistent with the highly anisotropic and non-convex structures observed in the final critic match loss landscape.}

\subsection{{Cross-system comparison}}

In this subsection, the quantitative analysis of the critical match loss landscape will be shown by comparing the landscapes' quantitative indexes of the two systems. The quantitative analysis of the loss landscape will also link with the system performance using the performance index in \autoref{subsection:system performance index}.

As shown in \autoref{fig:spc_3D loss}, the critic match loss landscape of the ADHDP algorithm for the spacecraft system is substantially more complex than that of the cart-pole system in \autoref{fig:cp_3D loss}. While the cart-pole case shows a smooth, single-slope surface, the spacecraft landscape contains multiple peaks and valleys, indicating a more challenging optimization process. \autoref{tab:cross-system quantative comparison} shows the quantitative description of the critic match landscapes using sharpness, basin area and local anisotropy. The normalized cost of the two system is also shown in the system, which enables the link between the characterisitcs of the loss landscapes and system performances.

\begin{table}[htbp]
\centering
\caption{{Comparison of ADHDP performance on different environments}}
\label{tab:cross-system quantative comparison}
\begin{tabular}{lcc}
\hline
 & \textbf{Cart-pole system} & \textbf{ S/C attitude system} \\
\hline
Control results   & stable   & unstable \\
Landscape shape   & Sloped   & non-convex \\
Sharpness $\mathrm{Sharp}_{\epsilon}$        & 7.952687 & 0.267753 \\
Basin area $A_{\rho}$            & 3.519303 & 20.596623 \\
Local anisotropy $\log\kappa$       & 1.445154 & 2.651535 \\
Normalized cost $\tilde J_H$   & 0.001029 & 0.148814 \\
\hline
\end{tabular}
\end{table}

For cart-pole, the controller converges. The landscape is essentially a single slope with a clear descent direction. \autoref{tab:cross-system quantative comparison} shows the quantitative indices of the loss landscape of cart-pole ADHDP control. Sharpness is relatively high, the basin area is small, and local anisotropy is low. This pattern is consistent with a monotone surface. The large sharpness indicates a stiff neighborhood along the slope around the final point. The small basin area $A_\rho$ shows a single tilted plane crossing the threshold only in a small region with low loss. The curvature around the final point is nearly isotropic with small $\log\kappa$. Under the same evaluation process, the system performance index is near zero, indicating successful control.

For the spacecraft (S/C) attitude case, the controller is unstable. The landscape presents multiple peaks and valleys and the path oscillates. Quantitatively, sharpness is small, the reported basin area is large, and local anisotropy is higher. The small sharpness reflects the lack of a single steep descent direction near the final point, while the large $A_\rho$ is caused by several shallow low-loss patches connected under the threshold in a non-convex surface, which does not imply robustness of the solution. The larger anisotropy indicates skewed curvature and narrow passages between valleys. If we combine the three indices and the observation of the loss landscape, it can be seen that, the algorithm terminates in a final point with a large and flat area yet with high local anisotropy, while the loss value is still high. Such a region makes it difficult for the optimization process to escape toward more favorable directions. The corresponding finite-horizon performance index is also larger, matching the observed instability.

From the analysis of the landscapes of the cart-pole and spacecraft attitude control system, it can be seen that the indices should therefore be interpreted in combination. Sharpness captures local stiffness. Basin area reflects the extent of low-loss regions but can inflate on fragmented landscapes. And anisotropy shows directional difficulty. Their joint pattern aligns with the system performance index.

The differences in the shape of the loss landscape are caused by the inherent dynamics of different systems. The number of states and the dimensions of the control input are different for the cart-pole system and the spacecraft attitude system. For the cart-pole system, the number of states is 4, which are the horizontal placement and angle placement, and their derivatives. The dimension of the control input is 1. For the combined spacecraft system, the number of states is 6, which are the three attitude angles and their derivatives. The dimension of the control input is 3, which is the control torques in three directions. The number of states and the dimensions of the control input align with the number of neurons at the input layer for the critic network and the actor network. The more neurons in the layer, the more complicated the structure of the neural networks, which means that more parameters need to be trained and updated.  As can be seen, with more states and dimensions of control, the structure of the neural network of the combined spacecraft system is more complicated than that of the cart-pole system. The differences in the loss landscape explicitly reveal the scale of difficulty in optimization of the J cost function and the training of the algorithm.

\autoref{fig:spc_2D_loss}  shows the 2-D projection of the 3-D critic match loss landscape. Different from \autoref{fig:cp_2D_loss}, of which the weight directly brings the loss from a larger value to a smaller value, the critic weight updates in the spacecraft attitude system exhibit a less stable pattern. The trajectory begins at a point associated with a relatively high critic loss and proceeds through a series of local minima, indicating frequent shifts in the optimization path rather than smooth convergence. During the journey of traveling between different local minima points, the critic loss value experiences spiky changes, which correspond with \autoref{fig:spc_critic loss}. 

However, it should be noted that the 2-D loss path does not strictly represent the critic loss at every iteration. The plot in \autoref{fig:spc_critic loss} provides the actual critic loss values recorded at each training step. In contrast, the critic match loss landscape is constructed by fixing a reference batch, namely the states from the final training episode and the corresponding temporal difference targets computed at the final episode, and then evaluating the critic match loss over candidate weights. Because online reinforcement learning evolves both data and targets over time, the landscape cannot reproduce every transient change. By tracking the weights and evaluating them on a fixed reference set, the landscape provides an interpretable view of the optimization geometry and a qualitative picture of how the algorithm evolves.

Considering the minor misalignment between the critic match loss landscape and critic loss curve with evolution, one naturally arrives at the question: why does the loss landscape still offer meaningful insight? 

Taking the cart-pole system as an example, the ADHDP control yields a convergent and successful result. The cost function is well approximated by the final critic, and the stored sequence of weights traces the optimization path toward a broad low region of the landscape. When the grid is evaluated on the final episode data and final episode targets, weights corresponding to earlier stages produce higher loss, while weights near the final solution produce lower loss, which appears as a darker region in the contour.

Let's recall the procedure of the generation of the critic match loss landscape. The weight grid is evaluated using data from the final training episode. In this context, weights located in the initial region of the grid correspond to earlier, undertrained stages of learning and therefore produce higher loss values. In contrast, weights in the later region represent well-trained stages that contribute to convergence, resulting in lower loss. Since the final episode is itself a converged simulation, the combination of well-trained weights and stable system response yields a low loss, which appears as a darker region in the landscape plot.

This alignment between the converged behavior and the corresponding low-loss region in the landscape demonstrates that the loss landscape captures the optimization trend, even if it does not reflect the exact loss values at each training step. When training is unstable, the landscape highlights directions where the cost approximation is sensitive or poor. The value of the critic match loss landscape lies not in representing precise numerical loss values, but in revealing key characteristics of the control problem and the evolution of the optimization process during training.

\subsection{Critic match loss landscape of final policy using random direction dim-reduction}
To examine whether the observed loss landscape characteristics are intrinsic to the optimization process rather than artefacts of a specific projection, an alternative dimensionality reduction is considered. While PCA provides a linear projection that emphasizes directions of large weight variation, it does not uniquely determine the
geometry of the loss surface. Therefore, a pair of random orthogonal directions is used as a complementary linear projection to assess the robustness of the landscape interpretation. It has to be mentioned that, only linear projections are considered here. Because the proposed loss landscape is constructed by explicitly reconstructing the critic network parameters on a two-dimensional subspace and re-evaluating the critic match loss. This requires a linear projection that admits an explicit inverse mapping from the reduced coordinates to the original parameter space. As a result, nonlinear embedding methods such as t-SNE therefore isn't considered, since they are less suitable 
 than linear dim-reduction methods for the reconstruction of valid network parameters for loss evaluation.

\begin{figure}[H]
    \centering
    \begin{subfigure}[b]{0.45\textwidth}
        \centering
        \includegraphics[width=\textwidth]{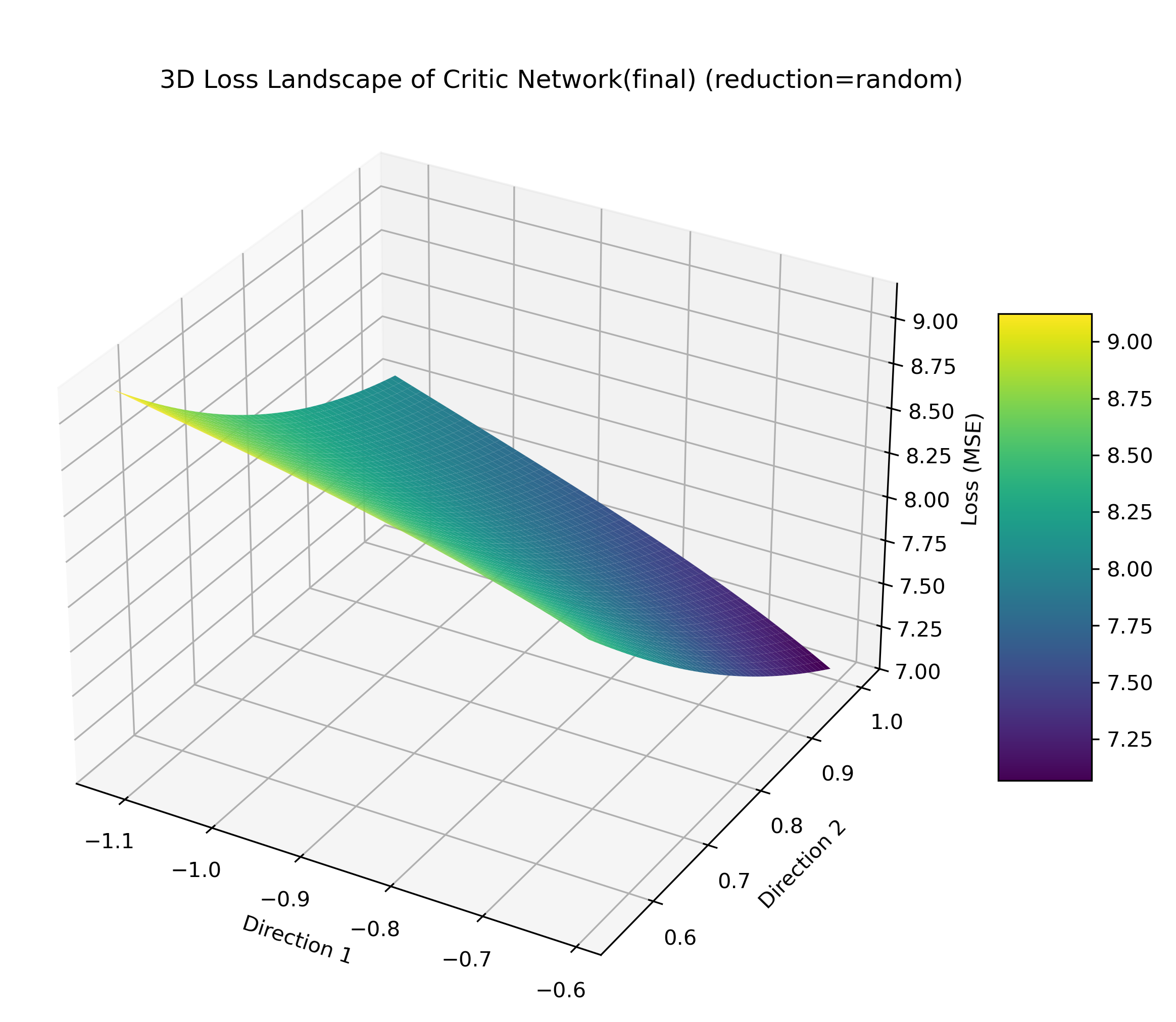}
        \caption{3-D loss of cart-pole ADHDP control}
        \label{fig:cp_3D loss random}
    \end{subfigure}
    \hspace{0.02\textwidth}
    \begin{subfigure}[b]{0.45\textwidth}
        \centering
        \includegraphics[width=\textwidth]{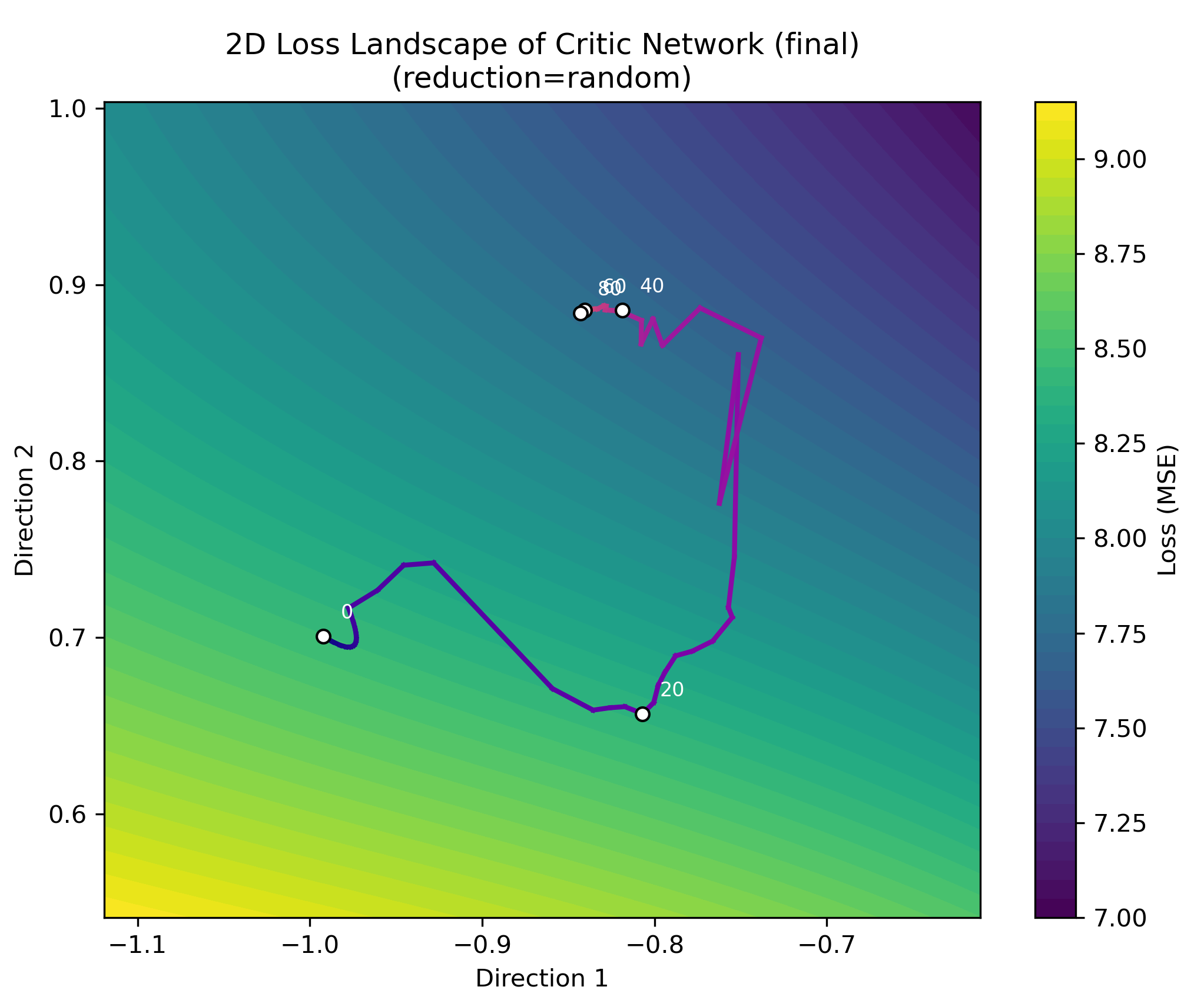}
        \caption{2-D loss path of cart-pole ADHDP control}
        \label{fig:cp_2D_loss random}
    \end{subfigure}
    \caption{3-D and 2-D loss landscape with random direction dim-reduction of cart-pole ADHDP control}
    \label{fig:cp_loss landscape random}
\end{figure}

\autoref{fig:cp_3D loss random} and \autoref{fig:cp_2D_loss random} show the 3-D surface and 2-D path when the weight space is projected onto two random orthogonal directions, instead of the PCA plane in \autoref{fig:cp_3D loss} and \autoref{fig:cp_2D_loss}. Compared with the PCA projection, the absolute loss range on the grid is compressed and the slope becomes less steep. Nevertheless, the overall terrain remains a single tilted surface. The 2-D optimization path still progresses almost monotonically along a dominant descent direction, with only minor turns near dense contour regions as it approaches the minimum.

This indicates that, although the numerical scale and apparent steepness depend on the choice of projection, the essential optimization characteristics remain unchanged. In particular, a clear descent channel is preserved on the projected plane under the final-policy reference, which is consistent with stable weight evolution and a low system performance index~$\tilde J_H$.

\begin{figure}[H]
    \centering
    \begin{subfigure}[b]{0.45\textwidth}
        \centering
        \includegraphics[width=\textwidth]{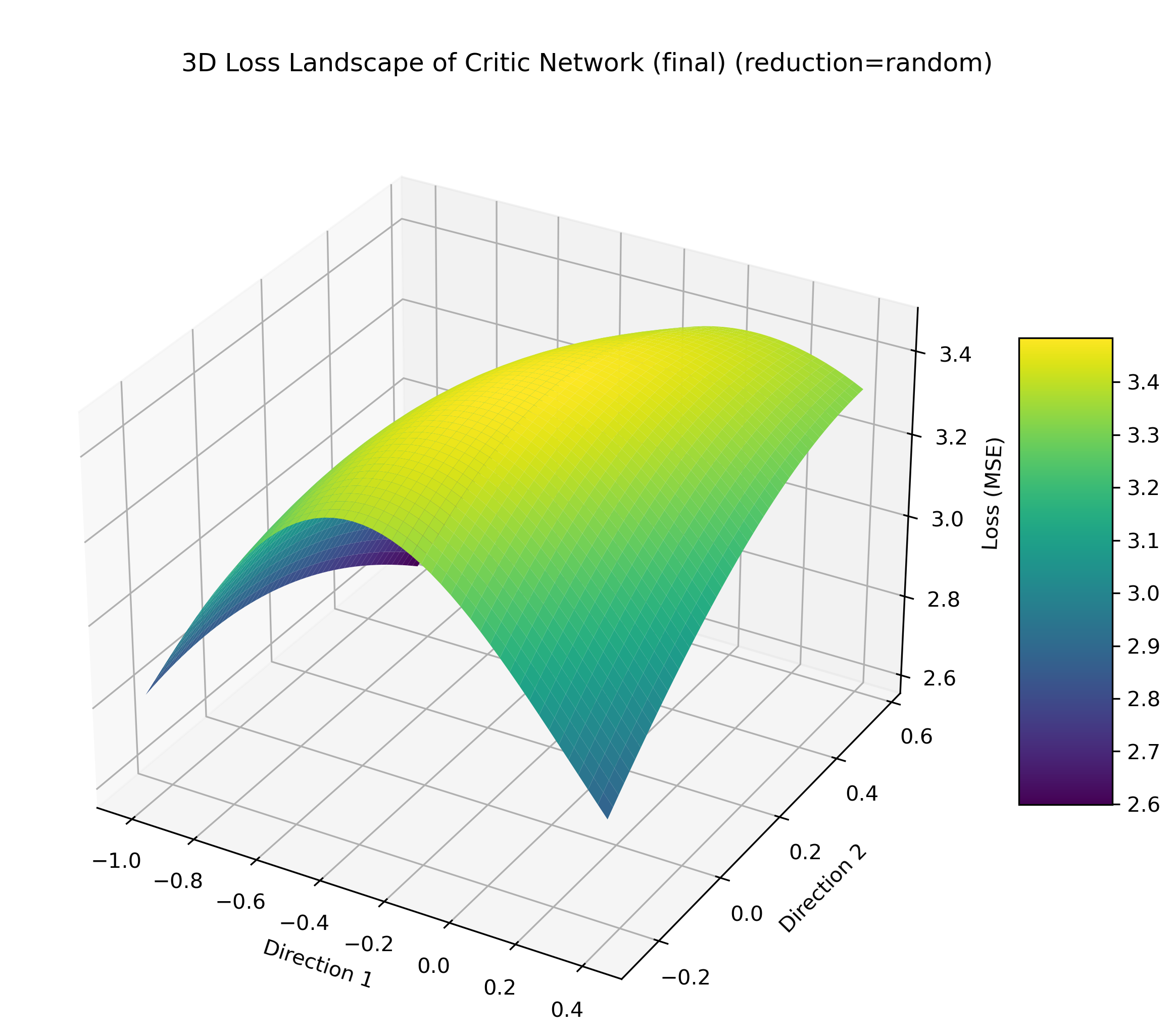}
        \caption{3-D loss landscape}
        \label{fig:spc_3D loss random}
    \end{subfigure}
    \hspace{0.02\textwidth}
    \begin{subfigure}[b]{0.45\textwidth}
        \centering
        \includegraphics[width=\textwidth]{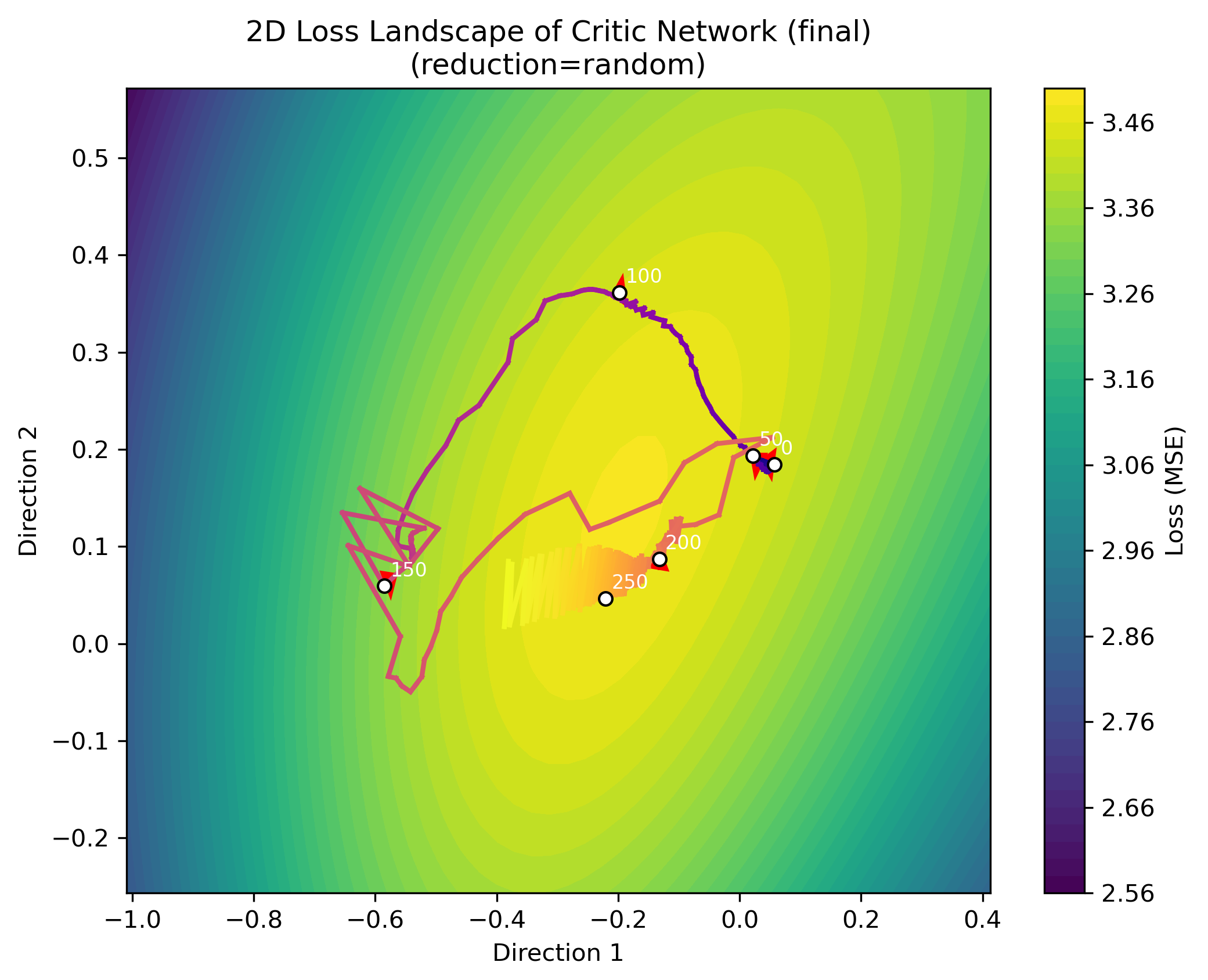}
        \caption{2-D loss landscape}
        \label{fig:spc_2D_loss random}
    \end{subfigure}
    \caption{3-D and 2-D loss landscape with random direction dim-reduction of spacecraft attitude ADHDP control}
    \label{fig:spc_loss landscape random}
\end{figure}
For spacecraft attitude control using ADHDP, \autoref{fig:spc_3D loss random} and \autoref{fig:spc_2D_loss random} show the 3-D surface and the 2-D path when the weight space is projected onto a pair of random orthogonal directions. As expected, the absolute loss range and the detailed surface morphology vary with the projection. Under the random projection, the surface appears as a shallow arch rather than a well-defined basin.  Along a ring near the top of this arch, the loss exhibits only weak variation, and the optimization path circulates around this region before settling, as shown in \autoref{fig:spc_2D_loss random}. This results in a visible loop on the 2-D contour, reflecting a small effective gradient in tangential directions and a lack of a strong descent direction.

From the case of the spacecraft attitude control, the interpretation of the algorithm behavior does not depend on using PCA or random orthogonal directions for dim-reduction. Both projections reveal a difficult and skewed landscape with limited descent, which is consistent with unstable training and poor control performance.

\subsection{Critic match loss landscape during training}
\label{subsection: loss landscape during training}
A full movie of the online evolution is not feasible, since both the reference data, states and TD targets, and the critic parameters change at every step, making the objective itself move continuously. While the two-dimensional optimization path provides a compact visualization of how the critic weights move during training, a loss landscape constructed only under the final policy cannot reflect the geometry perceived by the critic at earlier stages
of learning. To partially capture this temporal aspect, we fix the PCA plane built from the full sequence of episode-end critic weights and take mid-training snapshots. From the simulation setting in \autoref{section: ADHDP controlr result}, the mid-episode for cart-pole system is 50 and 150 for spacecraft attitude system. For each snapshot, the loss grid is re-centered at the episode-end weight of that episode. From \autoref{eq: contour Plots}, this re-centering means that the coordinate of each grid is calculated based on the distance between that grid and the selected episode-end weight. The critic match loss is evaluated on that plane using the states and TD targets from that same episode. This yields a mid-training landscape that is directly comparable with the final-policy landscape.
\begin{figure}[H]
    \centering
    \begin{subfigure}[b]{0.45\textwidth}
        \centering
        \includegraphics[width=\textwidth]{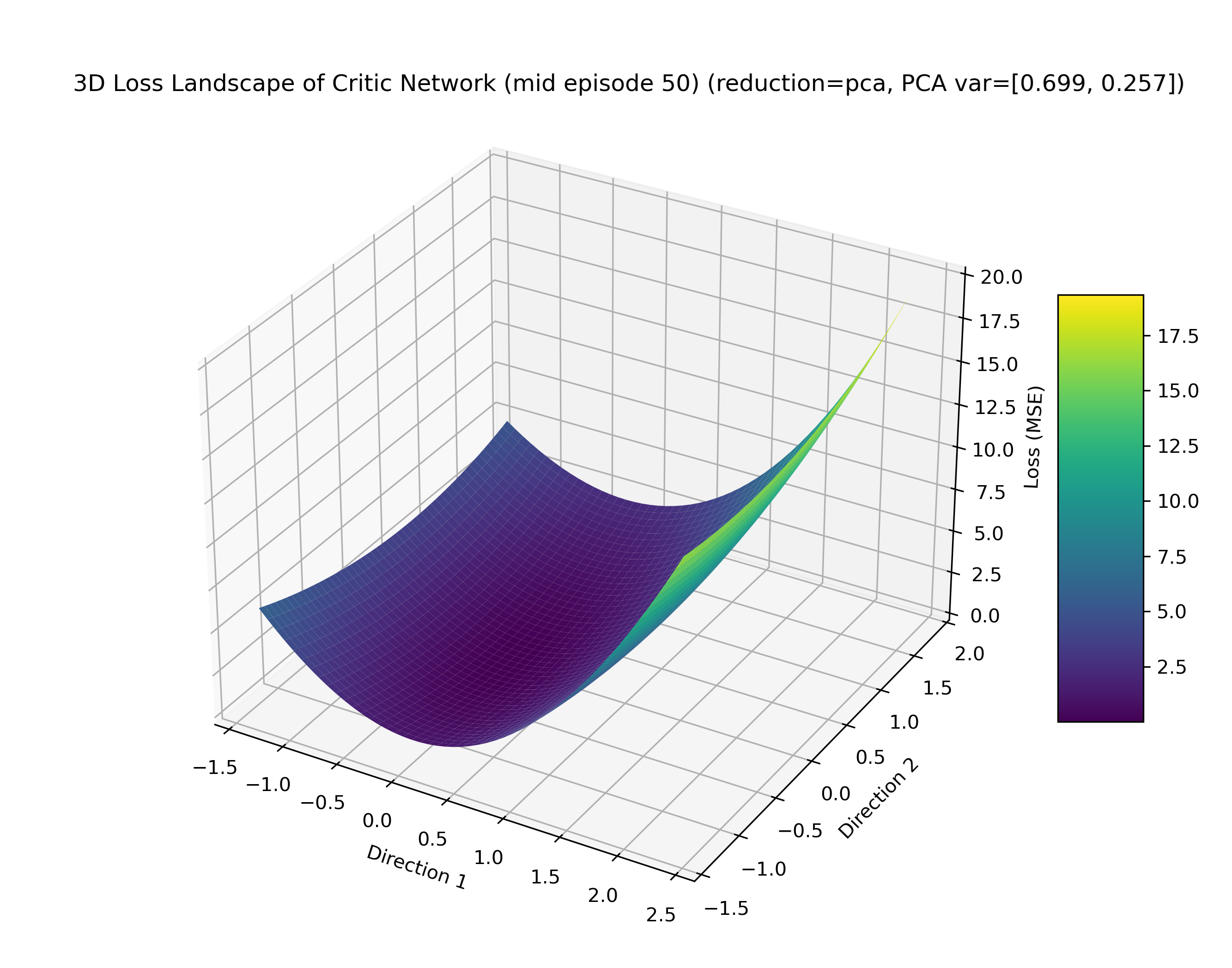}
        \caption{3-D loss of cart-pole ADHDP control during training}
        \label{fig:cp_3D loss mid}
    \end{subfigure}
    \hspace{0.02\textwidth}
    \begin{subfigure}[b]{0.45\textwidth}
        \centering
        \includegraphics[width=\textwidth]{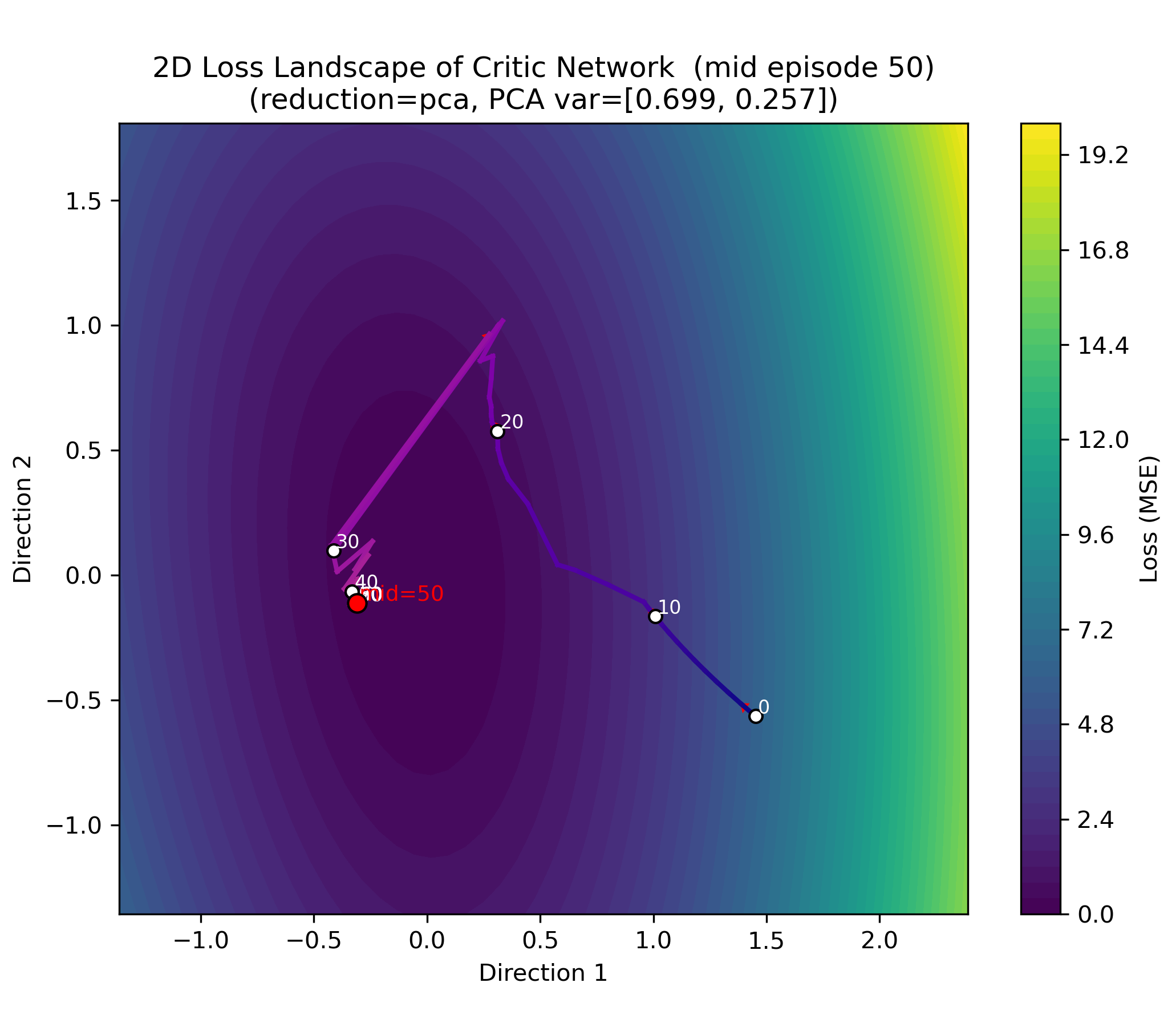}
        \caption{2-D loss path of cart-pole ADHDP control during training}
        \label{fig:cp_2D_loss mid}
    \end{subfigure}
    \caption{3-D and 2-D loss landscape cart-pole ADHDP control during training}
    \label{fig:cp_loss landscape mid}
\end{figure}
The mid-training surface as shown in \autoref{fig:cp_3D loss mid} and \autoref{fig:cp_2D_loss mid} already exhibits a coherent and nearly convex basin aligned with the descent direction. The two-dimensional optimization path progresses steadily toward the basin center, with only minor adjustments near regions of denser contours. Compared with the final-policy landscape shown in \autoref{fig:cp_3D loss}, the basin at episode 50 is shallower and slightly broader, indicating that the local curvature is still developing.

Nevertheless, the overall geometry is characterized by a single dominant slope, weak anisotropy, and the absence of competing valleys. This consistency between the mid-training and final landscapes explains the stable and monotonic convergence observed during training.

\begin{figure}[H]
    \centering
    \begin{subfigure}[b]{0.45\textwidth}
        \centering
        \includegraphics[width=\textwidth]{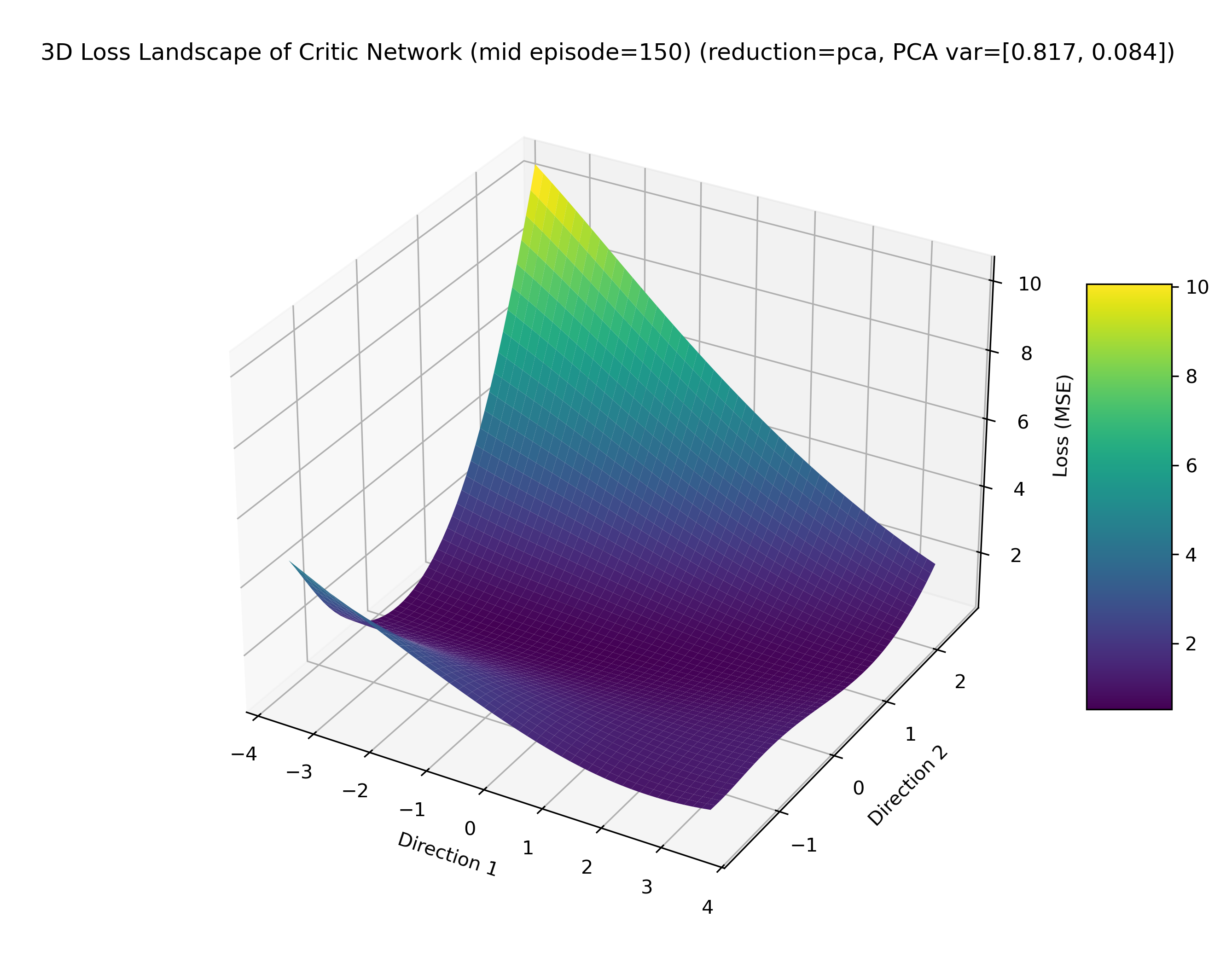}
        \caption{3-D loss of spacecraft attitude ADHDP control during training}
        \label{fig:spc_3D_loss_mid}
    \end{subfigure}
    \hspace{0.02\textwidth}
    \begin{subfigure}[b]{0.45\textwidth}
        \centering
        \includegraphics[width=\textwidth]{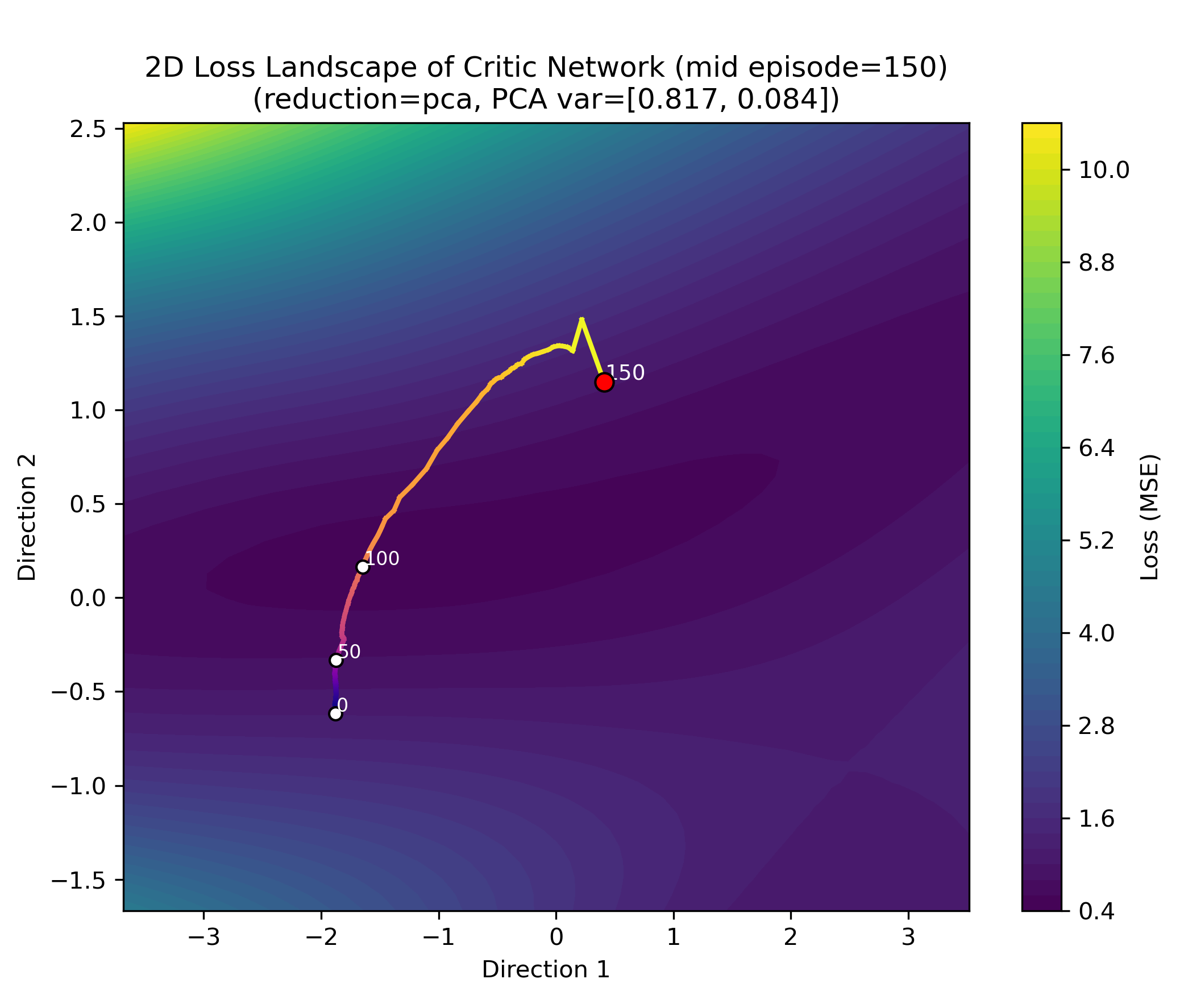}
        \caption{2-D loss path of spacecraft attitude ADHDP control during training}
        \label{fig:spc_2D_loss_mid}
    \end{subfigure}
    \caption{3-D and 2-D loss landscape spacecraft attitude ADHDP control during training}
    \label{fig:spc_loss_landscape_mid}
\end{figure}
The mid-training loss surface, shown in \autoref{fig:spc_3D_loss_mid} and \autoref{fig:spc_2D_loss_mid}, differs qualitatively from that of the cart-pole task. Around episode 150, the surface forms a shallow bowl-shaped region intersected by a ridge. Along this ridge, the two-dimensional optimization path turns and partially backtracks, before drifting through directions of low curvature. Compared with the final-policy landscape in \autoref{fig:spc_3D loss}, the mid-training surface is less cohesive, exhibiting weaker tangential curvature, while a narrow passage across the ridge remains visible. This shows that the critic optimization geometry is still evolving at this stage of training.

This behavior suggests a moving-target effect during learning. Gradients that are locally informative at early stages become misaligned with the geometry of the final basin. It brings step-size sensitivity and oscillatory motion near the ridge. As a consequence, system performance remains unstable and the normalized cost stays large. Overall, the temporal loss landscapes suggest that the instability arises not only from non-convexity, but also from the evolving learning signals provided by the critic. As these signals change over time, policy updates repeat traveling across low-cost regions with unstable features, preventing steady progress toward stable control.

\section{Conclusion}
{The critic loss landscape visualization method for online reinforcement learning is proposed to analyze the training behavior of actor–critic control algorithms. By constructing a critic match loss surface on a low-dimensional plane and overlaying the optimization path, the method provides an interpretable representation of how critic parameters evolve during critic learning. To analyze the resulting landscapes beyond visual inspection, quantitative indices and a normalized system performance index are introduced, allowing the geometric properties of the landscape to be related to control outcomes. The method is evaluated using the Action-Dependent Heuristic Dynamic Programming (ADHDP) algorithm on the cart-pole and spacecraft attitude control problems. Through comparisons across projection methods and training stages, the simulations reveal distinct landscape characteristics associated with stable convergence and unstable learning. Overall, the results show that the proposed visualization and analysis framework enables both qualitative and quantitative interpretation of critic optimization behavior in online reinforcement learning, offering a practical tool for interpreting and comparing actor–critic algorithms in system control applications.}

\appendix
\section{ADHDP Algorithm Details} 
\label{Appendix}
%% align or eqnarray environments can be used for multi line equations.
%% & is used to mark alignment points in equations.
%% \\ is used to end a row in a multiline equation.
The critic network and actor network are approximated using a MLP structure with one hidden layer.

In the critic network, the output $J(t)$ is in the form of ~\autoref{eq:cost function approx}. $w_{c_i}^{(2)}$ is the weight from the hidden layer to the output layer. 
\begin{eqnarray}
J(t)=\sum_{i=1}^{N_{h c}} w_{c_i}^{(2)}(t) p_i(t)
\label{eq:cost function approx}
\end{eqnarray}
\begin{eqnarray}
p_i(t)=\frac{1-e^{-q_i(t)}}{1+e^{-q_i(t)}}, i=1, \cdots, N_{h c}
\label{eq:actvfh}
\end{eqnarray}
\begin{eqnarray}
q_i(t)=\sum_{j=1}^{n+1} w_{c i j}^{(1)}(t) x_j(t), i=1, \cdots, N_{h c}
\label{eq:actvfo}
\end{eqnarray}

In the ~\autoref{eq:actvfh} and ~\autoref{eq:actvfo}, $q_i$ is the $i^{th}$ hidden node input of the critic network, $w_{c_i}^{(1)}$ is the weight from the input layer to the hidden layer and $p_i$  is the corresponding output of the hidden node. $N_{h c}$ is the total number of hidden nodes in the critic network. From  ~\autoref{eq:actvfh}, it can be seen that the hyperbolic-tangent function is used as the activation function at the hidden layer to add some nonlinearity to the training process. The linear function is used as the activation function at the output layer, as an example, as can be seen from ~\autoref{eq:cost function approx}. The choices for the activation function are decided according to the specific cases and users’ experience. 

Using a feed-forward network, the adaption in the action network is similar to the one in the critic network, while the inputs are the measured states, indicated with $x(t)$ in ~\autoref{fig:ADHDP_structure}, and the output is the action $u$. The associated equations for the action network are:
\begin{eqnarray}
u(t)=\frac{1-e^{-v(t)}}{1+e^{-v(t)}}
\label{eq:actfun_u}
\end{eqnarray}
\begin{eqnarray}
v(t)=\sum_{i=1}^{N_{h c}} w_{a_i}^{(2)}(t) g_i(t)
\label{eq:actfun_v}
\end{eqnarray}
\begin{eqnarray}
g_i(t)=\frac{1-e^{-h_i(t)}}{1+e^{-h_i(t)}}, i=1, \cdots, N_{h a}
\label{eq:actfun_g}
\end{eqnarray}
where $v(t)$ is the input to the action node, and $g_i$ and $h_i$ are the output and the input of the hidden nodes of the action network, respectively. From ~\autoref{eq:actfun_u}, it can be seen that the hyperbolic-tangent function is used as the activation function at the hidden layer. The hyperbolic tangent function is also used as the activation function at the output layer, as can be seen from ~\autoref{eq:actfun_g}. Similar to the critic network, the choices for the activation function are decided according to the specific cases and users’ experiences.
The training of the two neural networks is based on backpropagation. 
With the chain rule, the adaption of the critic network is summarized as follows.
From hidden to the output layer
\begin{eqnarray}
\Delta w_{c_i j}^{(2)}(t)=l_c(t)\left[-\frac{\partial E_c(t)}{\partial w_{c_i j}^{(2)}(t)}\right]
\label{eq:critic BBP1_O}
\end{eqnarray}
\begin{eqnarray}
\frac{\partial E_c(t)}{\partial w_{c_i j}^{(2)}(t)}=\frac{\partial E_c(t)}{\partial J(t)} \frac{\partial J(t)}{\partial w_{c_i j}^{(2)}(t)}=\gamma e_c(t) p_i(t)
\label{eq:critic BBP2_O}
\end{eqnarray}
where $\Delta w_{c_i}^{(2)}(t)$ indicates the weight from the hidden to the output layer of the critic network.
From the input to the hidden layer
\begin{eqnarray}
\Delta w_{c_{i j}}^{(1)}(t)=l_c(t)\left[-\frac{\partial E_c(t)}{\partial w_{c_{i j}}^{(1)}(t)}\right]
\label{eq:critic BBP1_i}
\end{eqnarray}
\begin{eqnarray}
\begin{aligned}
& \frac{\partial E_c(t)}{\partial w_{c_{i j}}^{(1)}(t)}=\frac{\partial E_c(t)}{\partial J(t)} \frac{\partial J(t)}{\partial p_i(t)} \frac{\partial p_i(t)}{\partial q_i(t)} \frac{\partial q_i(t)}{\partial w_{c_{i j}}^{(1)}(t)} \\
& =\gamma e_c(t) w_{c_i j}^{(2)}(t)\left[\frac{1}{2}\left(1-p_i^2(t)\right)\right] x_j(t) .
\end{aligned}
\label{eq:critic BBP2_i}
\end{eqnarray}
where $\Delta w_{c_i}^{(1)}(t)$ indicates the weight from the input to the hidden layer of the critic network.

The update rule for the nonlinear multi-layer action network also contains two sets of equations.

From the hidden to the output layer 
\begin{eqnarray}
\Delta w_{a_i}^{(2)}(t)=l_a(t)\left[-\frac{\partial E_a(t)}{\partial w_{a_i}^{(2)}(t)}\right]
\label{eq:actor_BBPO_1}
\end{eqnarray}
\begin{eqnarray}
\begin{aligned}
& \frac{\partial E_a(t)}{\partial w_{a_i}^{(2)}(t)}=\frac{\partial E_a(t)}{\partial J(t)} \frac{\partial J(t)}{\partial u(t)} \frac{\partial u(t)}{\partial v(t)} \frac{\partial v(t)}{\partial w_{a_i}^{(2)}(t)}\\
& =e_a(t) \sum_{i=1}^{N_{h c}}\left[w_{c_i}^{(2)}(t) \frac{1}{2}\left(1-p_i^2(t)\right) w_{c_{i, n+1}}^{(1)}(t)\right]\left[\frac{1}{2}\left(1-u^2(t)\right)\right] g_i(t)
\label{eq:actor_BBPO_2}
\end{aligned}
\end{eqnarray}
where $\Delta w_{a_i}^{(2)}$indicates the weight from the hidden to the output layer of the actor network.

In ~\autoref{eq:actor_BBPO_2}, ${\partial J(t)}/{\partial u(t)}$ is obtained from the critic network by changing variables and by the chain rule. The term $w_{c_{i, n+1}}^{(1)}$ is the weight associated with the input element from the action network output. Here, $i$ is the  $i^{th}$ node of the input layer, and $n$ is the number of nodes of the input layer. 

From the input to the hidden layer 
\begin{eqnarray}
\Delta w_{a_{i j}}^{(1)}(t)=l_a(t)\left[-\frac{\partial E_a(t)}{\partial w_{a_{i j}}^{(1)}(t)}\right]
\label{eq:actor_BBPi_1}
\end{eqnarray}
\begin{eqnarray}
\begin{aligned}
& \frac{\partial E_a(t)}{\partial w_{a_{i j}}^{(1)}(t)}=\frac{\partial E_a(t)}{\partial J(t)} \frac{\partial J(t)}{\partial u(t)} \frac{\partial u(t)}{\partial v(t)} \frac{\partial v(t)}{\partial g_i(t)} \frac{\partial g_i(t)}{\partial h_i(t)} \frac{\partial h_i(t)}{\partial w_{a_{i j}}^{(1)}(t)} \\
& =e_a(t) \sum_{i=1}^{N_{h c}}\left[w_{c_i}^{(2)}(t) \frac{1}{2}\left(1-p_i^2(t)\right) w_{c_{i, n+1}}^{(1)}(t)\right]\\
&\left[\frac{1}{2}\left(1-u^2(t)\right)\right] w_{a_i}^{(2)}(t)\left[\frac{1}{2}\left(1-g_i^2(t)\right)\right] x_j(t)
\end{aligned}
\label{eq:actor_BBPi_2}
\end{eqnarray}
where $w_{a_{i j}}^{(1)}(t)$ indicates the weight from the input to the hidden layer of the actor network.

In ADHDP implementations, ~\autoref{eq:critic BBP2_O} and ~\autoref{eq:critic BBP2_i} are used to update the weights in the critic network, while ~\autoref{eq:actor_BBPO_2} and ~\autoref{eq:actor_BBPi_2} are used to update the weights in the action network.

%%       \cite{<label>} ==> [1]

%%





% %% For numbered reference style

%   \textit{\LaTeX: a document preparation system},
%   Addison Wesley, Massachusetts,

% \end{thebibliography}

\bibliographystyle{unsrt} % We choose the "plain" reference style
\bibliography{main}

%\printbibliography

\end{document}